# BRF-GS: Hyperspectral Bidirectional Reflectance Factor Modeling and Image Generation Based on 3D Gaussian Splatting

*Yiling Yao[a, b, c], Wenjuan Zhang[a, c, *], Bowen Wang[a], Bocheng Li[a, c], Wentao Song[a, b, c], Bing Zhang[a, c]*

[a] *Aerospace Information Research Institute, Chinese Academy of Sciences, Beijing 100094, China*
[b] *International Research Center of Big Data for Sustainable Development Goals, Beijing 100094, China*
[c] *University of Chinese Academy of Sciences, Beijing 100049, China*

* *Corresponding author.*
*E-mail addresses: yaoyiling22@mails.ucas.ac.cn (Y. Yao), zhangbing@aircas.ac.cn (B. Zhang)*

**ABSTRACT**

The bidirectional reflectance factor (BRF) is a fundamental descriptor of the directional radiative properties of terrestrial surfaces. However, existing three-dimensional (3D) radiative transfer models rely on complex scene construction and computationally intensive radiative transfer solvers, making it challenging to efficiently generate multi-angle hyperspectral reflectance imagery. 3D Gaussian Splatting (3DGS) provides a new pathway for efficient neural scene representation and novel view synthesis, but its low-order spherical harmonics representation of view-dependent appearance is insufficient for capturing complex directional reflectance, while the high dimensionality of hyperspectral data and substantial differences in spectral quality across bands pose additional challenges to scene geometry reconstruction and spectral modeling. To address these challenges, we propose BRF-GS, a 3DGS-based framework for BRF modeling and reflectance image generation in hyperspectral remote sensing scenes. BRF-GS introduces a hybrid BRDF-driven kernel into 3DGS to improve the representation of complex directional reflectance responses, adaptively selects geometry-reliable bands to improve the reliability of 3D scene initialization, and employs a two-stage training strategy that decouples geometry optimization from spectral modeling to learn view-dependent reflectance from the full spectral range. We further construct and release the AIR-BRF dataset, a multi-angle hyperspectral directional reflectance dataset comprising three scenes covering diverse natural land-cover types and artificial targets. Experimental results demonstrate that BRF-GS achieves superior spatial fidelity and spectral accuracy in hyperspectral image generation, while accurately reproducing characteristic view-dependent BRF responses. This study provides an efficient data-driven approach for BRF modeling and multi-angle reflectance image generation in hyperspectral remote sensing scenes.

## 1. INTRODUCTION

Most terrestrial surfaces exhibit non-Lambertian reflectance characteristics, with reflectance varying substantially with illumination geometry, viewing geometry, and wavelength (Zhang et al., 2025). These directional reflectance properties arise primarily from the combined effects of 3D structure and material composition on the bidirectional distribution of reflected radiation (Nicodemus, 1965; Soenen et al., 2005). Characterizing the directional reflectance properties of terrestrial surfaces is fundamental to a range of applications, including quantitative retrieval of surface parameters (Román et al., 2011), investigation of radiative transfer mechanisms (Liu et al., 2021), land-cover and target identification (Xu et al., 2000) an so on. Multi-angle observations provide an essential basis for characterizing directional reflectance because they capture changes in surface reflectance under different observation geometries, enabling the modeling of bidirectional reflectance factors. Hyperspectral BRF imaging further provides fine-grained spectral information on directional reflectance, enabling more detailed characterization of wavelength-dependent reflectance behavior.

Despite the importance of multi-angle observations for characterizing directional reflectance, practical remote sensing systems generally provide only a limited number of viewing angles; for example, PROBA-1 provides hyperspectral observations from five viewing angles (Barnsley et al., 2004). Airborne and UAV-based multi-angle sensors further enable flexible acquisition of directional observations, but their angular sampling remains limited. The limited angular sampling of practical observations remains insufficient for fully characterizing the continuous directional reflectance. This limitation is particularly pronounced for targets with complex 3D structures, such as forest canopies and urban buildings, where directional reflectance can vary nonlinearly with viewing geometry. Sparse angular sampling therefore limits the comprehensive characterization of their directional reflectance behavior (Burkart et al., 2015; Stark et al., 2016).

To overcome the limitations of angular sampling in actual remote sensing observations, 3D radiative transfer models (RTM) have been widely used to simulate BRF images over a continuous range of viewing geometries, including DART (Gastellu-Etchegorry et al., 2017; Wang et al., 2022), DIRSIG (Schott et al., 1999; Ientilucci and Brown, 2003; Goodenough and Brown, 2017), and LESS (Qi et al., 2019). These approaches typically construct a 3D scene using triangular facets or volumetric elements, assign material and physical parameters to scene components, and simulate the interactions between radiation and scene elements using discrete radiative transfer solvers, Monte Carlo ray tracing, or related techniques. Although RTM provides an important physical approach with explicit physical meaning for simulating BRF images, it faces several limitations. First, scene construction and the specification of material and physical parameters generally require substantial prior knowledge and manual intervention, particularly for complex scenes. Second, the high cost of fine-grained 3D scene modeling and detailed land-cover classification often necessitates simplifications for complex targets, such as assigning shared physical properties to objects of the same class, thereby making it difficult to capture radiometric differences and fine-scale texture variations among objects within the same land-cover class, limiting the representation of spatially heterogeneous surface properties. Finally, the computational cost associated with geometric visibility determination and radiative transfer solution can be substantial (Qi et al., 2023).

3D Gaussian Splatting (3DGS) (Kerbl et al., 2023) has recently emerged as a representative approach for novel view synthesis in computer graphics. It represents a scene using a set of spatially distributed anisotropic Gaussian primitives and optimizes their positions, scales, opacities, and view-dependent appearance from multi-view images. With high-quality scene reconstruction and real-time rendering capabilities, 3DGS has become an important technique for novel view synthesis of natural scenes. Compared with conventional 3D radiative transfer models, 3DGS directly learns scene representations from multi-view observations, eliminating the need for manually constructed scenes and explicitly specified physical parameters. The parameters of individual Gaussian primitives can be optimized through backpropagation within a differentiable rendering framework, enabling the representation of spatially varying target appearance at fine spatial scales. Moreover, its rasterization-based rendering pipeline avoids the iterative radiative transfer solution required by RTM approaches. The original 3DGS achieves rendering rates exceeding 100 frames per second at 1080p resolution (Kerbl et al., 2023), enabling highly efficient novel view synthesis. These characteristics make 3DGS a promising framework for data-driven modeling of directional reflectance and BRF image generation.

However, hyperspectral BRF images in remote sensing differ substantially from multi-view natural images in both data characteristics and reconstruction requirements. First, remote sensing scenes typically cover large spatial extents and contain diverse targets with distinct 3D structures and different directional reflectance characteristics. The low-order spherical harmonics (SH) representation adopted by the original 3DGS primarily captures smooth, low-frequency variations in view-dependent appearance and therefore has limited capability to represent strongly directional and nonlinear reflectance responses (Yang et al., 2024). Second, hyperspectral BRF images contain hundreds of spectrally contiguous bands, and each pixel represents a physically meaningful reflectance spectrum rather than merely visual appearance. For quantitative remote sensing, accurate reconstruction of the spectral shape and magnitude across the full spectral range is therefore essential. Directly extending RGB-based representations to hundreds of spectral channels would substantially increase the number of model parameters and computational burden, while treating spectral channels independently would fail to exploit strong inter-band correlations and may compromise spectral fidelity. Third, the data quality of hyperspectral observations can vary substantially across spectral bands, resulting in pronounced inter-band differences in signal quality. Such quality variations may cause some bands to provide unreliable information for feature extraction and geometric reconstruction, potentially affecting geometric initialization and introducing noise-related artifacts in the generated images. These characteristics make existing 3DGS frameworks unsuitable for direct application to hyperspectral BRF modeling and reflectance image generation.

Although systematic studies of 3DGS for remote sensing BRF modeling and publicly available datasets remain limited, several studies in natural image rendering have explored related challenges: to address the limited capability of spherical harmonics to represent complex directional appearance, methods such as GaussianShader (Jiang et al., 2024), Ref-Gaussian (Yao et al., 2025), and Spec-Gaussian (Yang et al., 2024) incorporate microfacet-based bidirectional reflectance distribution function (BRDF) models, such as the Cook–Torrance model, to improve the representation of specular reflection. Microfacet-based BRDF models are primarily designed to describe local surface reflectance by statistically modeling the distribution of microscopic surface normals, such as specular highlights. However, their local surface reflectance formulation does not explicitly account for the broader interactions between directional reflectance and complex 3D scene structure. In remote sensing, directional reflectance can be jointly influenced by the 3D arrangement of scene elements, occlusion and shadowing, inter-object interactions, and multiple scattering, particularly in structurally complex environments. These complex scene-level effects cannot be adequately captured by a single microfacet-based BRDF model, limiting its ability to comprehensively represent directional reflectance in remote sensing scenes.

To address the high dimensionality and spectral fidelity challenges of hyperspectral imagery, HyperGS (Thirgood et al., 2025) and Hyperspectral Gaussian Splatting (Narayanan et al., 2025) extend the 3DGS framework to hyperspectral representations. Hyperspectral BRF modeling requires high spectral fidelity because each pixel represents a physically meaningful reflectance spectrum, while the large number of spectral bands imposes substantial computational and storage demands. HyperGS projects hyperspectral images into a low-dimensional latent space through an autoencoder to reduce computational and storage costs, whereas Hyperspectral Gaussian Splatting introduces wavelength encoding to obtain a compact spectral representation across the full spectral range. Although these strategies improve the efficiency of hyperspectral representation, latent compression may suppress subtle spectral variations and narrow absorption features, potentially compromising physically meaningful spectral signatures. Moreover, both approaches jointly optimize spectral representations and spatial geometry, which may introduce geometry–reflectance coupling and optimization ambiguity.

To alleviate the effects of varying data quality across hyperspectral bands, HyperGS extracts features from all spectral bands and introduces depth-aware densification strategies, while Hyperspectral Gaussian Splatting incorporates a hyperspectral diffusion model for image denoising. However, these approaches primarily focus on mitigating the effects of noisy or low-quality observations rather than explicitly assessing the reliability of individual spectral bands for geometric reconstruction. Given the substantial inter-band differences in signal quality, treating all bands equally may cause unreliable bands to contribute to feature extraction and geometric initialization, potentially compromising the reconstructed geometry and introducing artifacts into the generated images. A more effective strategy should therefore distinguish the geometric reliability of different spectral bands and selectively exploit bands that provide more reliable information for geometric reconstruction. Thus, the potential of exploiting band-dependent data quality for reliable geometric reconstruction in remote sensing hyperspectral BRF modeling remains insufficiently explored.

Moreover, these methods have primarily been developed and evaluated on natural-scene hyperspectral datasets, and their applicability to remote sensing scenes with complex 3D structures and directional reflectance characteristics remains unexplored.

To address these challenges and the scarcity of remote sensing hyperspectral BRF datasets, we propose BRF-GS, a 3D Gaussian Splatting framework for BRF modeling and hyperspectral directional reflectance image generation in remote sensing scenes. The main contributions are summarized as follows:

- **A geometry-reliable spectral bands selection strategy for hyperspectral Gaussian initialization.** We develop a geometry-reliable spectral bands selection strategy that jointly considers band-wise signal quality and spatial feature responses. An equivalent feature count (EFC) is introduced to quantify band-wise geometric reliability, enabling adaptive selection of geometry-informative bands for robust 3D geometry initialization and subsequent hyperspectral Gaussian optimization.
- **A hybrid BRDF-driven kernel representation for Gaussian primitives.** We introduce a hybrid semi-empirical BRDF kernel-driven representation into Gaussian primitives, integrating Lambertian diffuse reflection, a simplified microfacet-based specular component, volumetric scattering, and geometric-optical effects. Learnable weights adaptively estimate the contributions of different radiative mechanisms, enabling a compact and physically interpretable representation of complex directional reflectance across diverse remote sensing targets.
- **A geometry-spectral decoupled two-stage training framework.** By

exploiting the view-invariant property of geometry and the view-dependent nature of directional reflectance, we first optimize geometric parameters using selected geometry-reliable bands and subsequently introduce full hyperspectral observations to learn spectral directional reflectance variations.

- **The AIR-BRF dataset.** We establish AIR-BRF, a multi-angle hyperspectral BRF images dataset acquired by an unmanned aerial vehicle (UAV) over three remote sensing scenes, including diverse natural land-cover types and artificial targets. The dataset provides a benchmark for BRF image generation at remote sensing scales and systematic evaluation of hyperspectral 3DGS methods.

The remainder of this paper is organized as follows. Section 2 reviews related work. Section 3 presents the proposed BRF-GS method in detail. Section 4 reports the experimental results and analysis. Section 5 concludes the paper and discusses future research directions.

## 2. RELATED WORK

### *2.1. 3D Gaussian Splatting and Directional Reflectance Modeling*

3D Gaussian Splatting (3DGS) (Kerbl et al., 2023) was introduced as a novel explicit scene representation method that models scene geometry and appearance using a set of parameterized 3D Gaussian primitives. The geometry of each Gaussian primitive is described by its 3D center position $\boldsymbol{\mu} \in \mathrm{R}^3$ and covariance matrix $\boldsymbol{\Sigma} \in \mathrm{R}^{3\times 3}$, which jointly determine its spatial location and shape. In terms of appearance representation, the original 3DGS models view-dependent color using spherical harmonics up to the third order and separately optimizes opacity $\alpha \in [0,1]$. Spherical harmonics are a set of orthogonal basis functions defined on the unit sphere $S^2$. The $l$ order SH basis consists of $2l+1$ basis functions, and the color $c_k$ of the Gaussian $k$ is represented as an SH expansion conditioned on the viewing direction $d$:

$$c_k(d)=\sum_{l=0}^{3}\sum_{m=-l}^{l}c_{k,l}^{m}\cdot Y_l^m(d), \tag{1}$$

where $Y_l^m$ is the $l$-th-order $m$-th SH basis function and $c_{k,l}^m$ are the corresponding learnable coefficients. During rendering, the 3D Gaussian primitives are projected onto the 2D image plane through affine transformation, sorted by depth, and alpha-blended to obtain the color of pixel $x$:

$$C(x)=\sum_{i\in N}c_i(d_i)\alpha_i\prod_{j=1}^{i-1}(1-\alpha_j), \tag{2}$$

However, the SH-based appearance representation of the original 3DGS has inherent limitations in modeling complex anisotropic appearance: low-order SH expansions have limited capability in capturing high-frequency directional variations, while the number of SH coefficients increases quadratically with the SH order. For hyperspectral data containing hundreds of spectral channels, extending SH-based representations to the spectral domain would introduce a substantial number of additional parameters for each Gaussian primitive, resulting in significant computational and memory overhead.

To this end, several studies have improved directional reflectance modeling in the natural RGB image domain. GaussianShader (Jiang et al., 2024) introduced learnable surface normals and replaced the implicit SH-based directional appearance modeling with a physically based shading function consisting of diffuse, specular, and residual components. This enables Gaussian primitives to encode local surface orientation and significantly improves reconstruction quality for scenes containing reflective surfaces. Spec-Gaussian (Yang et al., 2024) replaced SH with anisotropic spherical Gaussians (ASGs), leveraging their intrinsic suitability for representing narrow-lobe high-frequency signals to better capture anisotropic specular reflection details. 3DGS-DR (Ye et al., 2024) adopted a deferred shading framework to decouple geometry and illumination for improved reconstruction of specular surfaces. Building upon this idea, Ref-GS (Zhang et al., 2025a) exploited the more accurate normal representation of 2DGS (Huang et al., 2024), introduced illumination-direction factorization, and incorporated roughness-aware deferred rendering to achieve high-quality reconstruction and reflective rendering of smooth objects with specular reflection and self-reflection. MaterialRefGS (Zhang et al., 2025b) further constrained Gaussian representations to maintain view-consistent physical attributes and introduced photometric consistency modeling, improving novel view synthesis robustness and physical parameter estimation in highly reflective scenes. Collectively, these methods have advanced 3DGS beyond pure geometric reconstruction and appearance fitting toward more physically grounded directional reflectance modeling.

However, the above methods share a common limitation in their application scale: they mainly focus on close-range RGB images of individual objects, such as metallic utensils and glass products, and employ microfacet BRDF models such as Cook–Torrance (Cook and Torrance, 1982) for reflection modeling. These models characterize the statistical properties of surface micro-geometry through parameters such as roughness and describe the distribution of local surface orientations using microfacet normal distribution functions such as GGX (Walter et al., 2007). Although effective for directional reflectance modeling dominated by surface microstructure, such models are insufficient for remote sensing observations, where scenes typically consist of heterogeneous natural and artificial targets with diverse material properties. At remote sensing scales, the directional reflectance response of ground objects is governed not only by intrinsic material characteristics but also by multi-scale interactions among component mixing, 3D structural occlusion, shadowing effects, and multiple scattering. Therefore, a single microfacet BRDF model cannot adequately describe these complex radiative mechanisms, motivating the development of modeling frameworks that integrate multiple physical processes, including volumetric scattering and geometric-optical effects. Furthermore, existing approaches are primarily developed in the image domain; while they can reconstruct view-dependent appearance variations in 2D image space, they do not explicitly model the wavelength-dependent spectral characteristics of directional reflectance.

### *2.2. 3D Gaussian Splatting for Spectral Imaging*

With the advantages demonstrated by 3DGS in RGB-image-based 3D reconstruction and novel view synthesis, recent studies have extended Gaussian representations from RGB imagery to multispectral and hyperspectral imagery, aiming to jointly model spectral properties and 3D scene geometry.

For multispectral imagery, SpectralGaussians (Sinha et al., 2025) was among the first attempts to replace the RGB appearance attributes in 3DGS with multispectral reflectance. It further introduced a spectral shading model incorporating diffuse and specular reflection components to replace the original spherical harmonics appearance representation, enabling spectral appearance modeling and assigning semantic features to Gaussian primitives for 3D reconstruction and semantic analysis of spectral scenes. However, this method still models spectral channels independently and does not explicitly exploit the intrinsic correlations among contiguous spectral bands. Moreover, it was mainly evaluated on small-scale objects and synthetic multispectral datasets, leaving its applicability to complex remote sensing scenes with diverse targets and heterogeneous spectral responses insufficiently investigated.

To overcome the limitations of independent spectral parameterization and improve spectral compactness, MS-Splatting (Meyer et al., 2026) proposed a unified neural spectral representation that learns a low-dimensional feature vector for each Gaussian and employs a multilayer perceptron (MLP) to decode the view-dependent full-spectrum response. This strategy exploits the intrinsic correlations among spectral bands, thereby reducing storage requirements and computational complexity. However, the learned spectral representation mainly relies on neural appearance fitting without explicitly incorporating physical priors of surface reflectance and illumination–viewing geometry. Although effective for compact spectral representation, it remains challenging to model complex directional reflectance caused by surface structures and anisotropic reflectance mechanisms in remote sensing scenarios.

For hyperspectral imagery, HyperGS (Thirgood et al., 2025) extended 3DGS to hyperspectral scenes by addressing the high dimensionality and computational complexity associated with hyperspectral representations. It employs a convolutional autoencoder to project high-dimensional spectral information into a compact latent space, optimizes latent spectral features associated with Gaussian primitives, and reconstructs full spectra through spectral decoding. Moreover, HyperGS adopts grayscale-based geometric initialization and introduces a depth-aware gradient densification strategy to alleviate unstable Gaussian growth caused by hyperspectral noise. A combination of Charbonnier loss and cosine similarity loss is further utilized to improve spectral reconstruction fidelity. However, the latent-space compression may discard subtle spectral variations, particularly narrow absorption features and high-frequency spectral details, when the latent representation is insufficiently expressive. Furthermore, grayscale-based geometric initialization does not explicitly exploit the complementary spatial structures and spectral discriminability among different hyperspectral bands, which may limit robust geometry recovery in complex remote sensing scenarios.

Hyperspectral Gaussian Splatting (Narayanan et al., 2025) further alleviates the need for explicit spectral compression by introducing a wavelength encoder that embeds wavelength information into a learnable feature space and employs an MLP to predict wavelength-dependent offsets of spherical harmonic coefficients, enabling full-spectrum Gaussian representation. It further incorporates spectral constraints based on KL divergence and cosine similarity to preserve spectral distribution consistency and employs a hyperspectral diffusion model for spectral noise reduction. However, this framework still jointly optimizes geometric parameters and high-dimensional spectral radiometric parameters, leading to potential geometry–reflectance appearance entanglement during optimization. Moreover, its noise-handling strategy mainly focuses on improving spectral reconstruction quality, while overlooking the varying reliability of different spectral bands for 3D geometric reconstruction.

Furthermore, existing hyperspectral 3DGS approaches have been predominantly validated on the close-range single-plant observation dataset introduced in HS-NeRF (Ma and He, 2024). Comprehensive evaluations on remote sensing hyperspectral scenes with heterogeneous surface types and multi-angle observations remain limited. Consequently, existing spectral 3DGS methods have not yet fully addressed the challenges of hyperspectral remote sensing applications, including high-dimensional spectral representation, reliable geometry reconstruction from hyperspectral data, and directional reflectance modeling.

### *2.3. Directional Reflectance Modeling Methods in Remote Sensing*

At the remote sensing observation scale, surface directional reflectance is commonly described using the BRF and BRDF (Nicodemus et al., 1977; Schaepman-Strub et al., 2006). Its variation with illumination and viewing geometry is determined by coupled effects of surface reflection, volume scattering, and 3D geometric structure. Based on radiative transfer and geometric-optical theories, various physically based models have been developed to describe directional reflectance characteristics, including the SAIL model (Verhoef, 1984), the Li–Strahler geometric-optical model (Li and Strahler, 1985), and the four-scale geometric-optical model (Chen and Leblanc, 1997). Building upon these physical models, semi-empirical kernel-driven BRDF models (Roujean et al., 1992; Wanner et al., 1995; Lucht et al., 2000) were further proposed. These approaches represent directional reflectance as a linear combination of physically interpretable kernels, where each kernel represents a specific physical component of directional reflectance. Since kernel functions are primarily determined by illumination–viewing geometry and only a limited number of weighting parameters require estimation, kernel-driven models achieve a favorable balance between physical interpretability, parameter efficiency, and computational effectiveness. Consequently, they have been widely adopted for generating global surface albedo and nadir-adjusted reflectance products, such as MODIS MCD43 (Schaaf et al., 2002).

However, conventional kernel-driven models mainly characterize directional reflectance at the pixel level, where complex 3D structural effects are not explicitly represented but implicitly embedded in model parameters. Consequently, they cannot explicitly describe fine-scale spatial structures, visibility relationships, and local directional reflectance variations induced by heterogeneous surface geometry. This limitation restricts their capability to generate multi-angle reflectance products with explicit 3D structural information. Recently, several studies have incorporated topographic effects into kernel-driven models, including KDST (Wu et al., 2018), TCKD (Hao et al., 2020), and Topo-KD (Yan et al., 2020), to improve directional reflectance estimation over rugged terrains. Nevertheless, these approaches still rely on pixel-level parameterization and remain limited by conventional BRDF formulations in representing complex 3D scenes.

An alternative category of RTM approaches is based on 3D radiative transfer modeling. These methods explicitly represent 3D scene structures and simulate electromagnetic radiation transport and interactions using radiative transfer solvers, Monte Carlo ray tracing, and related numerical techniques, thereby enabling physically based BRF simulation. Representative models include DART (Gastellu-Etchegorry et al., 2017; Wang et al., 2022), DIRSIG (Schott et al., 1999; Ientilucci and Brown, 2003; Goodenough and Brown, 2017), and LESS (Qi et al., 2019). Compared with kernel-driven BRDF models, 3D radiative transfer models provide a more explicit representation of 3D geometry, shadowing and occlusion effects, and multiple scattering processes, offering stronger physical interpretability for analyzing directional reflectance mechanisms in complex scenes.

However, RTM approaches typically require explicit 3D scene representations, such as voxel-based structures, polygonal models, or other geometric descriptions, together with detailed assignments of geometric, material, and optical properties to individual scene elements. The construction of such physically based scenes requires substantial scene-specific modeling and prior knowledge, limiting their scalability for automated reconstruction of real-world remote sensing scenes. Moreover, acquiring accurate surface optical properties at fine spatial scales remains challenging. Consequently, practical applications often simplify heterogeneous surfaces by assigning identical optical properties to surfaces within the same land-cover or material class, which may overlook fine-scale variability in material properties and local directional reflectance characteristics. In addition, the computation of complex visibility relationships, shadowing, and radiative transfer processes remains computationally intensive, hindering efficient simulation and reconstruction of high-spatial-resolution multi-angle remote sensing reflectance. Therefore, developing an efficient framework that can directly exploit real observations to model 3D directional reflectance while preserving fine and complex directional reflection mechanisms remains a critical challenge in remote sensing BRF modeling.

## 3. METHOD

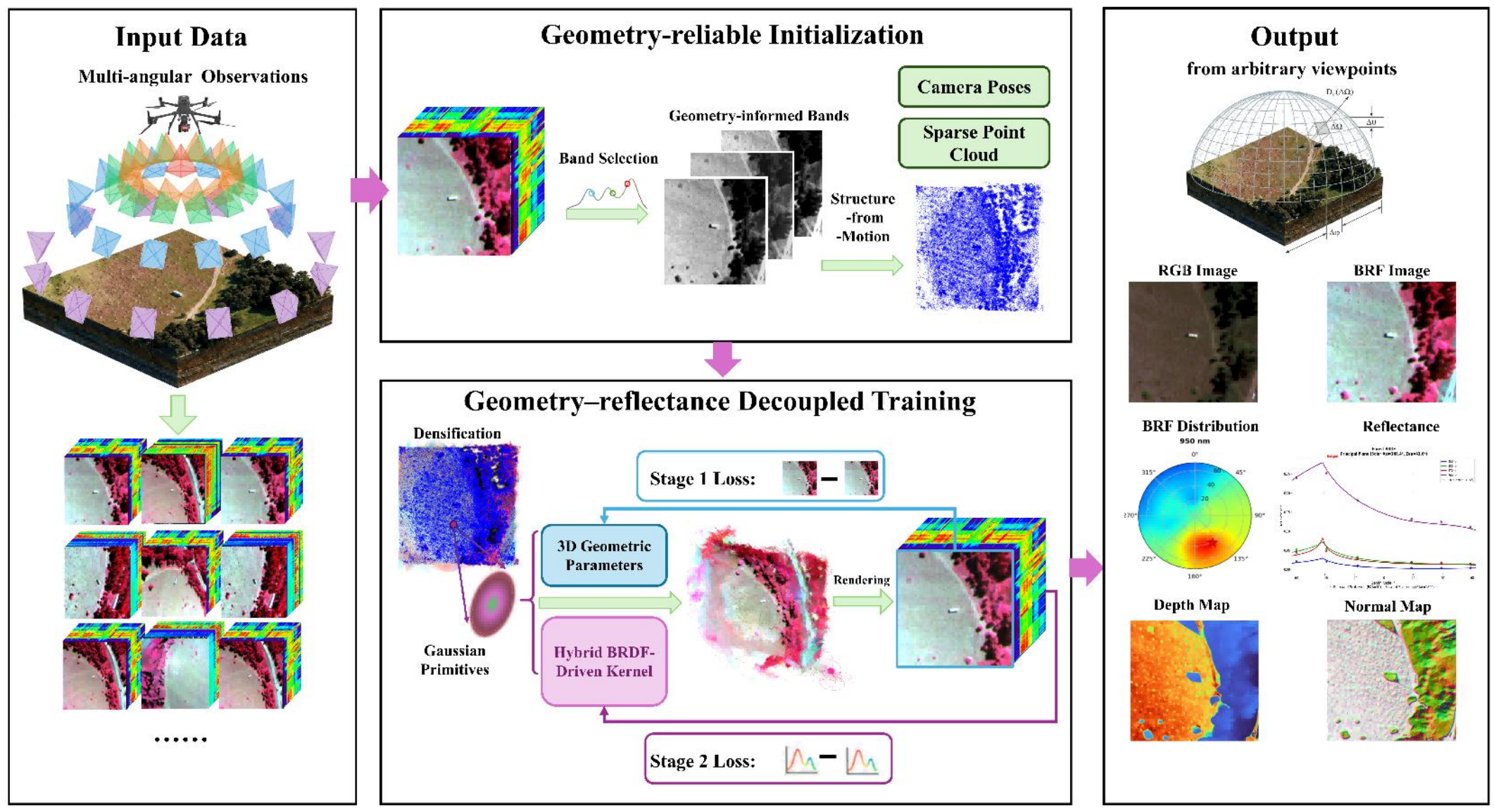

**Fig. 1** Overview of the proposed BRF-GS framework. Multi-angle hyperspectral observations are first used to identify geometry-reliable spectral bands for robust Structure-from-Motion initialization. A geometry-reflectance decoupled two-stage optimization strategy is subsequently applied to estimate scene geometry and spectral directional reflectance. The optimized Gaussian representation enables arbitrary-view rendering and directional reflectance analysis, including BRF/BRDF modeling.

This section presents the proposed BRF-GS framework. As illustrated in Fig. 1, the proposed method takes multi-angle hyperspectral reflectance observations as input and aims to generate hyperspectral images, directional reflectance responses (BRF/BRDF), depth maps, and normal maps under arbitrary viewing directions. The framework consists of three key components: (1) geometry-reliable spectral bands selection and initialization (Section 3.1); (2) hybrid BRDF-driven Gaussian kernel optimization (Section 3.2); and (3) a geometry-reflectance decoupled two-stage training framework (Section 3.3).

### *3.1. Geometry-reliable Spectral Bands Selection Strategy for Hyperspectral Gaussian Initialization.*

This subsection presents a geometry-reliable spectral bands selection strategy for hyperspectral Gaussian initialization. In the standard 3DGS framework, Gaussian primitives are typically initialized using Structure-from-Motion (SfM), which extracts local features from multi-view images, estimates camera poses, and reconstructs a sparse 3D point cloud. However, hyperspectral remote sensing imagery contains hundreds of contiguous spectral bands, whose suitability for geometric reconstruction varies substantially due to differences in sensor response characteristics, signal quality, and spatial feature quality. Some spectral bands are more susceptible to noise and exhibit insufficient spatial texture, resulting in unreliable feature detection and matching, whereas bands with higher-quality and more distinctive spatial features can provide more robust geometric constraints for camera pose estimation and 3D structure recovery. Directly evaluating the geometric suitability of hyperspectral bands through inter-image feature matching is computationally expensive, especially for hyperspectral datasets containing hundreds of spectral bands and multi-view observations. Moreover, rigorous band-wise SNR characterization requires sensor-specific radiometric calibration and noise modeling, which are not always available in practical remote sensing applications. Therefore, instead of performing exhaustive spectral matching or precise radiometric quality estimation, we introduce a computationally efficient geometry-reliable spectral bands selection strategy and retain only geometrically informative bands with high signal quality and reliable spatial feature responses for Gaussian geometric initialization and subsequent geometry parameter optimization.

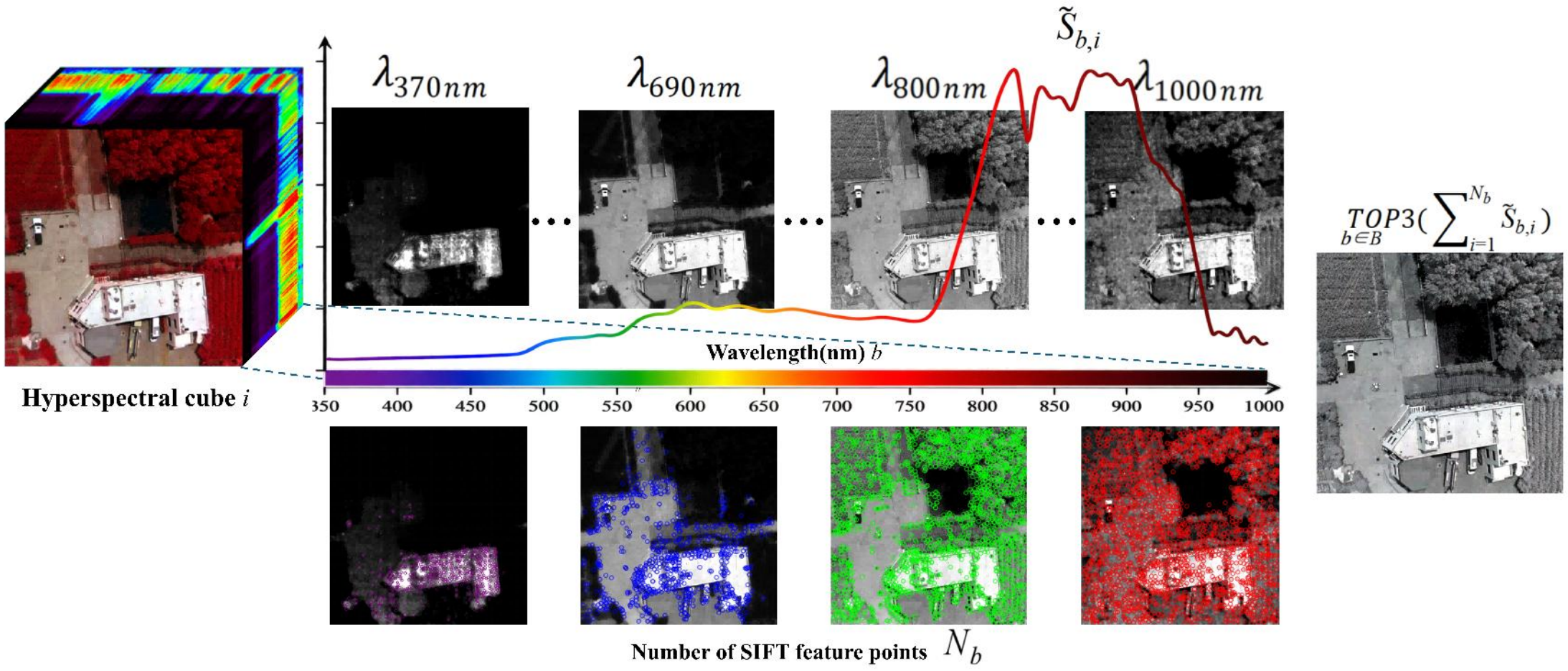

Fig. 2 Illustration of the geometry-reliable spectral band selection strategy for hyperspectral geometric initialization. The hyperspectral cube is evaluated band-by-band to identify the spectral band with the highest geometric reliability. As example illustrated, the bands at 370 nm and 690 nm provide insufficient detectable features, while the 1000 nm band produces abundant but unreliable features caused by noise. In contrast, the band near 800 nm exhibits strong signal response and generates stable spatial features, making it suitable for feature extraction and geometric recovery.

Specifically, given a hyperspectral reflectance image containing $B$ spectral bands acquired under a specific observation geometry, we first perform local feature extraction independently on each band image treated as a grayscale image. For the $b$-th spectral band, the detected feature-point set is denoted as

$$F_b = \{f_{b,i}\}_{i=1}^{N_b}, \quad \textbf{(3)}$$

Where $N_b$ represents the number of detected feature points in the $b$-th band.

The noise level of each band is estimated using the black-cover calibration measurements acquired during data collection. For the $b$-th band, the standard deviation of the black-cover response, denoted as $\sigma_b^{\text{dark}}$, is used to approximate the background noise level under negligible incident radiance. For each detected feature point $f_{b,i}$, a 5×5 local neighborhood window $N(f_{b,i})$ is constructed, and the median reflectance within this window is adopted as the local effective signal intensity:

$$I_{b,i} = Median\{R_b(x,y) \mid (x,y) \in N(f_{b,i})\}, \quad \textbf{(4)}$$

where $R_b(x,y)$ represents the reflectance value of the $b$-th band at pixel $(x,y)$. A relative signal quality coefficient is then defined as:

$$S_{b,i} = \frac{I_{b,i}}{\sigma_b^{\text{dark}} + \epsilon}, \quad \textbf{(5)}$$

where $\epsilon$ is a small positive constant introduced to avoid numerical instability when $\sigma_b^{\text{dark}}$ is close to zero.

To enable comparison among different spectral bands within the same scene, the relative signal quality coefficients are normalized as:

$$\tilde{S}_{b,i} = \frac{S_{b,i}}{\max\limits_{j \in \{1,\dots,B\}} \max\limits_{k \in \{1,\dots,N_j\}} S_{j,k}}, \quad \textbf{(6)}$$

resulting in normalized values within the range of [0,1]. The equivalent feature count (EFC) of the $b$-th band is subsequently calculated by accumulating the signal quality weighted feature responses:

$$EFC_b = \sum_{i=1}^{N_b} \tilde{S}_{b,i}, \quad \textbf{(7)}$$

All spectral bands are ranked based on their EFC values, and the three bands with the highest geometric reliability are selected:

$$\lambda_{geo} = \underset{b \in B}{TOP3}\,(EFC_b)\,, \tag{8}$$

$\lambda_{geo}$ is explicitly retained as the geometric representation of the hyperspectral observation, while the remaining spectral information is implicitly encoded into a compact representation through a 1D convolutional autoencoder. Specifically, the original $B$-dimensional reflectance vector $\mathbf{r} \in \mathbb{R}^B$ is mapped into a $d$-dimensional vector $\mathbf{z} \in \mathbb{R}^d (d \ll B)$:

$$\mathbf{z} = E_\phi(\mathbf{r}) : \mathbb{R}^B \rightarrow \mathbb{R}^d, \tag{9}$$

The selected bands $\lambda_{geo}$ are used for feature extraction, multi-view feature matching, camera pose estimation, and sparse 3D point cloud initialization. The resulting camera parameters and initial sparse 3D point cloud provide initialization for subsequent hyperspectral Gaussian optimization.

### *3.2. Hybrid BRDF-driven Kernel Representation for Gaussian Primitives*

This section introduces a BRDF-driven kernel representation to replace the low-order spherical harmonics used in conventional 3DGS, enabling view-dependent BRF modeling for hyperspectral remote sensing. At the remote sensing observation scale, the BRF contributions result from the combined effects of multiple scattering mechanisms, including Lambertian diffuse scattering, volumetric scattering, geometric-optical effects, and surface specular reflection. Inspired by semi-empirical kernel-driven BRDF models, which represent BRF as a weighted combination of predefined kernels, and further incorporating microfacet reflection theory, we model the wavelength-dependent BRF characteristics of each Gaussian primitive as the superposition of four physically meaningful components: an isotropic diffuse term, a volumetric scattering term, a geometric-optical term, and a microfacet specular term.

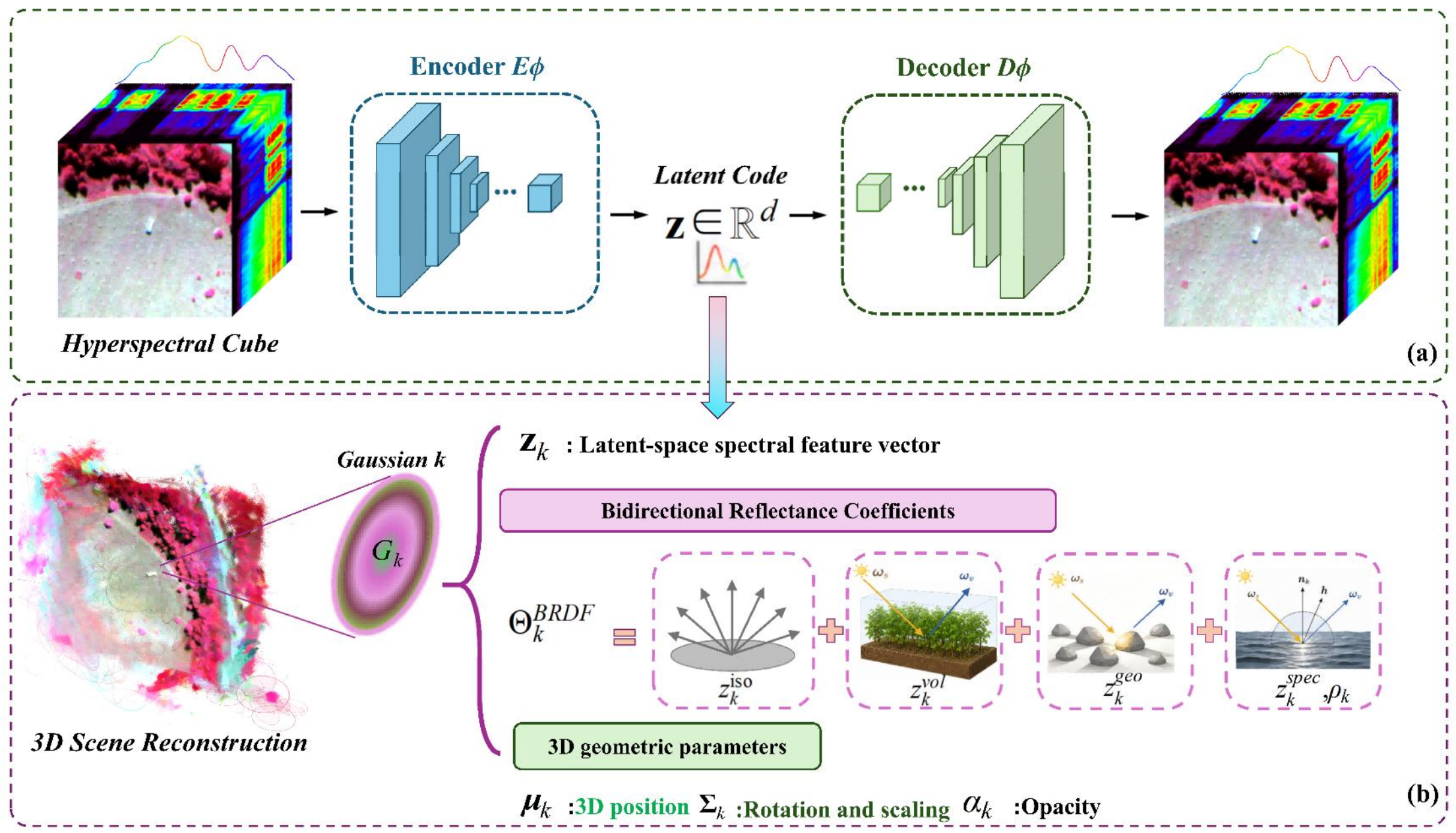

Fig. 3 Illustration of the hybrid BRDF-driven kernel representation for Gaussian primitives. (a) Spectral representation learning using an encoder–decoder network, where the hyperspectral cube is compressed into compact spectral feature vectors assigned to Gaussian primitives and reconstructed through the spectral decoder.(b) Representation of a BRDF-aware Gaussian primitive, which consists of geometry parameters, a spectral features vector, and hybrid BRDF kernel parameters modeling multiple directional reflectance mechanisms.

We define each 3D Gaussian as a 3D reflectance primitive that jointly encodes spatial geometric properties and directional reflectance characteristics:

$$G_k=\{\boldsymbol{\mu}_k,\boldsymbol{\Sigma}_k,o_k,\mathbf{z}_k,\Theta_k^{BRDF}\}, \tag{10}$$

where $\boldsymbol{\mu}_k$ denotes the Gaussian center, $\boldsymbol{\Sigma}_k$ represents the covariance matrix describing its spatial extent and orientation , $o_k$ is the opacity parameter, $\mathbf{z}_k$ is the spectral feature vector for hyperspectral compression, and $\Theta_k^{BRF}$ controls the directional reflectance response:

$$\Theta_k^{BRDF}=\left\{z_k^{\text{iso}},z_k^{vol},z_k^{geo},z_k^{spec},\rho_k\right\}, \tag{11}$$

where $z_k^{\mathrm{iso}}, z_k^{vol}, z_k^{geo}$ and $z_k^{spec}$ represent latent spectral coefficients associated with different scattering mechanisms, and $\rho_k$ denotes the learnable surface roughness parameter.

For the Gaussian primitive $k$, the directional spectral reflectance at wavelength $\lambda$, illumination direction $\omega_i$, and viewing direction $\omega_o$ is first modeled in the compact spectral space:

$$z_k(\omega_i,\omega_o)=z_k^{iso}+z_k^{vol}K_{Ross}(\omega_i,\omega_o)+z_k^{geo}K_{Li}(\omega_i,\omega_o)+z_k^{spec}K_{Cook}(\omega_i,\omega_o,\mathrm{n}_k,\rho_k), \quad \textbf{(12)}$$

where $K_{Ross}$, $K_{Li}$, and $K_{Cook}$ represent the volumetric scattering, geometric-optical, and microfacet specular reflection kernels, respectively. The wavelength-dependent reflectance spectrum is subsequently reconstructed through the spectral decoder:

$$r_k(\lambda,\omega_i,\omega_o)=D_\phi\big(z_k(\omega_i,\omega_o)\big), \quad \textbf{(13)}$$

where $D_\phi$ denotes the learned spectral decoder that maps the compact latent representation into the continuous hyperspectral reflectance domain.

***Volumetric scattering component***

The volumetric scattering contribution is modeled using the RossThick kernel (Roujean et al., 1992):

$$K_{\mathrm{Ross}}=\frac{(\pi/2\text{-}\xi)\cos\xi+\sin\xi}{\cos\omega_i+\cos\omega_o}-\frac{\pi}{4}, \quad \textbf{(14)}$$

where $\xi$ is the phase angle:

$$\cos\xi=\cos\omega_i\cos\omega_o+\sin\omega_i\sin\omega_o\cos\phi, \quad \textbf{(15)}$$

and $\phi$ denotes the relative azimuth angle between illumination and observation directions.

***Geometric-optical component***

The geometric-optical scattering component adopts the Li-Sparse kernel (Wanner et al., 1995):

$$K_{\mathrm{Li}}=O-\sec\omega_i-\sec\omega_o+\frac{1+\cos\xi}{2\cos\omega_i\cos\omega_o}, \quad \textbf{(16)}$$

where

$$O=\frac{1}{\pi}[t-\sin t\cos t]\left(\frac{1}{\cos\omega_i}+\frac{1}{\cos\omega_o}\right), \quad \textbf{(17)}$$

and

$$t=\arccos\left(\frac{1}{b}\sqrt{D^2+(\tan\omega_i\tan\omega_o\sin\phi)^2}\right), \quad \textbf{(18)}$$

$$D=\sqrt{\tan^2\omega_i+\tan^2\omega_o-2\tan\omega_i\tan\omega_o\cos\phi}, \quad \textbf{(19)}$$

***Microfacet specular component***

To characterize directional reflection from smooth surfaces such as water surfaces and metallic objects, the specular component is modeled using the Cook–Torrance (Cook and Torrance, 1982) microfacet BRDF with the GGX normal distribution function (Walter et al., 2007):

$$K_{Cook}=\frac{D_{GGX}(h,n_k,\rho_k)\cdot F(\omega_o,h)}{4(n_k\cdot\omega_o)(n_k\cdot\omega_i)}, \quad \textbf{(20)}$$

where $h$ is the half-vector:

$$h=\frac{\omega_i+\omega_o}{\|\omega_i+\omega_o\|}, \quad \textbf{(21)}$$

$n_k$ denotes the surface normal estimated from the Gaussian covariance matrix, and $\rho_k$ is the learnable roughness parameter.

The GGX normal distribution function is defined as:

$$D_{GGX}=\frac{\rho_k^2}{\pi\left[\left(n_k\cdot\ \mathrm{h}\right)^2\left(\rho_k^2-1\right)+1\right]^2}, \quad \textbf{(22)}$$

and the Fresnel term follows the Schlick approximation:

$$F(\omega_o,\mathrm{h})=F_0+(1-F_0)\left(1-\mathrm{h}\cdot\ \omega_o\right)^5. \quad \textbf{(23)}$$

The Fresnel term is approximated using the Schlick model (Schlick, 1994) with a fixed normal-incidence reflectance $F_0=0.04$, following the common assumption for dielectric surfaces. Material-dependent spectral variations of specular reflection are implicitly captured by the learned latent specular component $z_k^{spec}$.

### *3.3. Geometry–Reflectance Decoupled Two-Stage Gaussian Optimization Strategy*

This subsection proposes a geometry–reflectance decoupled two-stage Gaussian optimization strategy, which decomposes hyperspectral Gaussian optimization into two sequential stages: geometry reconstruction from reliable spectral observations and directional spectral reflectance learning. This strategy is motivated by the observation that jointly optimizing geometry and spectral reflectance from the initial stage may introduce ambiguity between geometric and reflectance parameters. This is because geometric structures are wavelength-independent, whereas directional reflectance variations are strongly wavelength- and view-dependent.

In the first stage, the geometrically reliable spectral bands $\lambda_{geo}$ selected in Section 3.1 are used to reconstruct a stable 3D geometric representation. Specifically, the sparse point cloud generated by SfM is used to initialize 3D Gaussian primitives with geometric parameters and band-specific reflectance parameters. The geometric parameters are optimized together with the reflectance coefficients of the selected band.

The optimization objective of the first stage is defined as:

$$L_{stage1}=(1-\alpha)L_1+\alpha L_{D-SSIM}, \tag{24}$$

where $L_1$ denotes the pixel-wise loss calculated on the reflectance values of the selected geometry-reliable spectral bands, and $L_{D-SSIM}$ represents the structural similarity loss computed on its corresponding grayscale images. The weighting factor $\alpha$ balances the contributions of the two loss terms.

After obtaining a stable geometric reconstruction, the second stage focuses on learning the directional spectral reflectance characteristic. The geometry parameters $\boldsymbol{\mu}_k$, $\boldsymbol{\Sigma}_k$, and $o_k$, and spectral feature vector $Z_k$ are fixed, while only the bidirectional reflectance coefficients $\Theta_k^{BRDF}$ are optimized. In this stage, all available hyperspectral bands are introduced to reconstruct the continuous spectral reflectance response under different viewing directions.

To simultaneously constrain spectral shape consistency and absolute reflectance accuracy, the second stage employs a combination of reflectance reconstruction loss and spectral cosine similarity loss:

$$L_{stage2}=(1-\beta)L_{refl}+\beta L_{cos}, \tag{25}$$

where

$$L_{refl}=\frac{1}{|\Omega|B}\sum_{x\in\Omega}\sum_{b=1}^{B}|R(x,\lambda_b)-\widehat{R}(x,\lambda_b)|, \tag{26}$$

and

$$L_{cos}=\frac{1}{|\Omega|}\sum_{x\in\Omega}\left(1-\frac{\widehat{R}_x^T R_x}{\|\widehat{R}_x\|_2\|R_x\|_2}\right). \tag{27}$$

Here, $R_x$ and $\widehat{R}_x$ denote the ground truth and reconstructed hyperspectral reflectance vectors of pixel $x$, respectively; $B$ is the number of spectral bands; $\Omega$ represents the set of pixels involved in optimization.

## 4. EXPERIMENTS AND ANALYSIS

To evaluate the proposed BRF-GS framework under diverse remote sensing scenes, we constructed the AIR-BRF dataset.

### *4.1. AIR-BRF Dataset Construction*

#### *4.1.1 Study Area Selection*

To systematically evaluate the capability of BRF-GS for modeling directional reflectance across remote sensing scenes with diverse land-cover types and their corresponding BRF characteristics, three representative areas were selected to construct the AIR-BRF dataset: the HuaiLai Remote Sensing Test Site, the SaiHanBa Mechanical Forest Farm, and the GuanTing Reservoir in Hebei Province.

The test site scene covers diverse land-cover types, including broadleaf forest, corn field, pond, building, cement surface, and vehicles, representing a complex scene composed of both natural and artificial surfaces with diverse directional reflectance characteristics. The forest farm scene contains typical natural land-cover types, including coniferous forest, dry grasslands, shrubs, and haystacks, exhibiting distinct anisotropic reflectance behaviors associated with vegetation. The reservoir scene includes freshwater, bare soil, rubble, grasslands, and wetlands, capturing diverse directional reflectance variations among different natural surface materials.

The RGB images of the three scenes and their corresponding representative land-cover types are shown in **Fig.** 4.

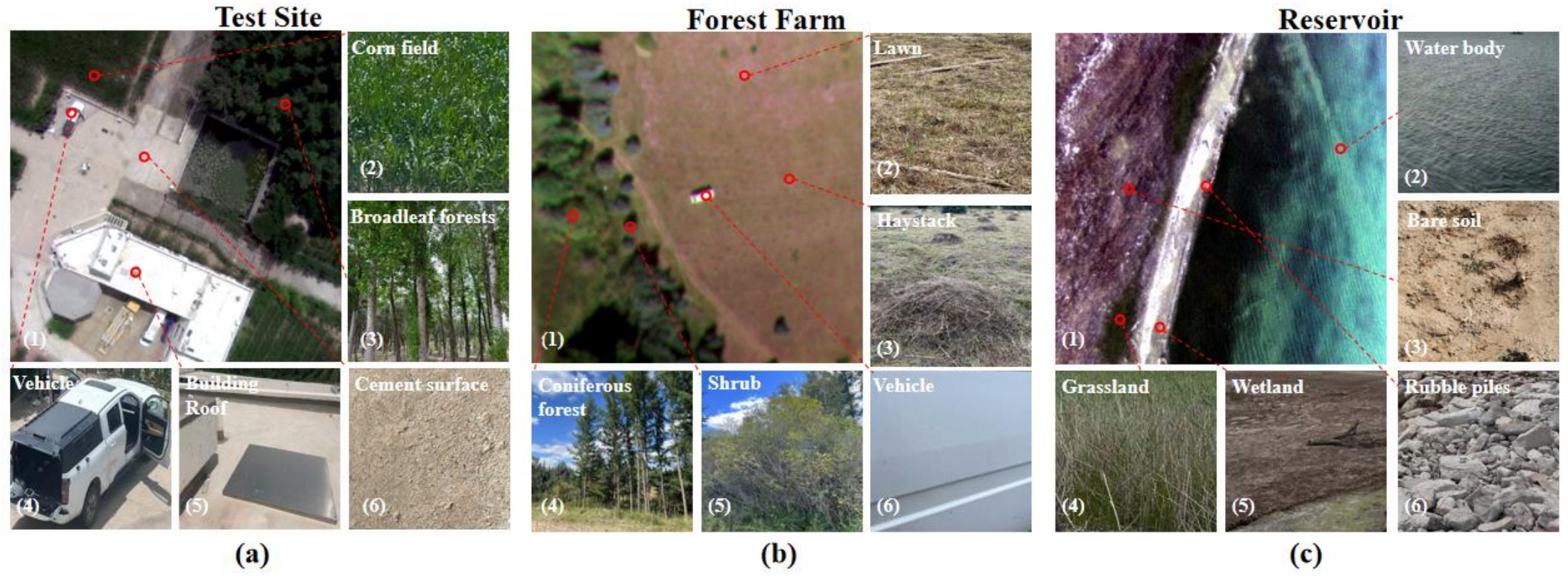

**Fig.** 4 Optical RGB images of the three scenes and their corresponding representative land-cover types. (a) Test site; (b) Forest farm; (c) Reservoir.

### *4.1.2 Data Acquisition*

Multi-angle hyperspectral data were acquired using a DJI M300 RTK UAV equipped with a Cubert Ultris X20P hyperspectral imaging system, as shown in Fig. 5(a). The X20P covers a spectral range of 350 – 1000 nm with a spectral resolution of 4 nm, providing 164 spectral bands. It employs a snapshot hyperspectral imaging architecture. Each image contains 1886×1886 pixels, with a spatial resolution of approximately 10 – 20 cm/pixel depending on flight altitude and observation geometry, and a field of view (FOV) of 35°.

The three scenes were surveyed during two acquisition campaigns in June and September. For each scene, multi-angle observations were collected from 12 azimuth directions at 30° intervals (0°, 30°, 60°, 90°, 120°, 150°, 180°, 210°, 240°, 270°, 300°, and 330°) and five representative zenith angles (0°, 18°, 23°, 38°, and 58°). A total of 49 hyperspectral images were acquired for each scene, resulting in 147 images across the three scenes. To minimize changes in solar illumination geometry during the multi-angle acquisition, the observations for each scene were completed within approximately 15 min. All data were collected around midday under clear-sky and low-wind conditions to maintain relatively stable illumination conditions. The solar position during each acquisition campaign is indicated by the red stars in Fig. 5, while the blue markers indicate the viewing directions and the yellow dashed lines denote the UAV flight trajectories.

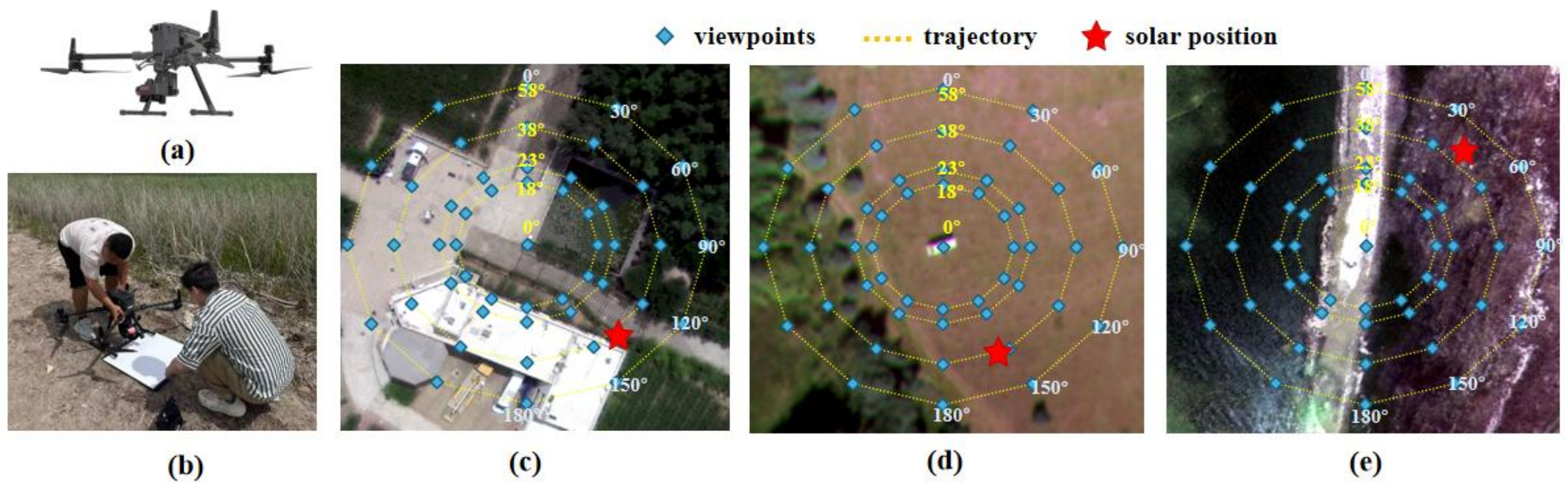

**Fig**. 5 Data acquisition and scene composition of the AIR-BRF dataset. (a) Data acquisition system. (b) Radiometric calibration with a standard white reference panel. (c–e) RGB images of the Test Site, Forest Farm, and Reservoir scenes, respectively. Blue markers indicate the viewing directions of acquisition, red stars indicate corresponding solar positions, and yellow dashed lines denote the UAV flight trajectories.

### *4.1.3 Data Preprocessing*

Radiometric calibration was performed using a standard white reference panel with known reflectance (Fig. 5b), converting the raw digital numbers (DNs) into calibrated reflectance. A black reference target was additionally used to estimate the sensor dark current and noise level. Because the X20P uses a snapshot hyperspectral imaging architecture, spatial offsets may occur among spectral bands. Inter-band registration was therefore performed to spatially align the spectral bands, achieving a registration accuracy of approximately 0.1 pixels. The calibrated and registered hyperspectral images were then used for subsequent BRF-GS reconstruction and evaluation. The images were randomly divided into training and testing sets at a ratio of 6:1.

### *4.2. BRF Image Generation Comparison*

Based on the AIR-BRF dataset constructed above, we selected representative state-of-the-art methods with publicly available implementations for comparative evaluation. The compared methods include the original 3DGS (Kerbl et al., 2023), GaussianShader (Jiang et al., 2024), Ref-GS (Zhang et al., 2025a), and MS-Splatting (Meyer et al., 2026). The first three are RGB-image-based methods, which are designed for three-channel image inputs and therefore cannot directly reconstruct hyperspectral reflectance. To enable a consistent comparison, they were trained and evaluated using color-infrared (CIR) composite images generated from the hyperspectral observations. MS-Splatting is a spectral-image-based method and was trained and evaluated using the corresponding spectral images. All methods used the same training and testing sets. Qualitative and quantitative comparisons were conducted to evaluate their capability for novel view BRF image generation.

### *4.2.1 Qualitative Comparison*

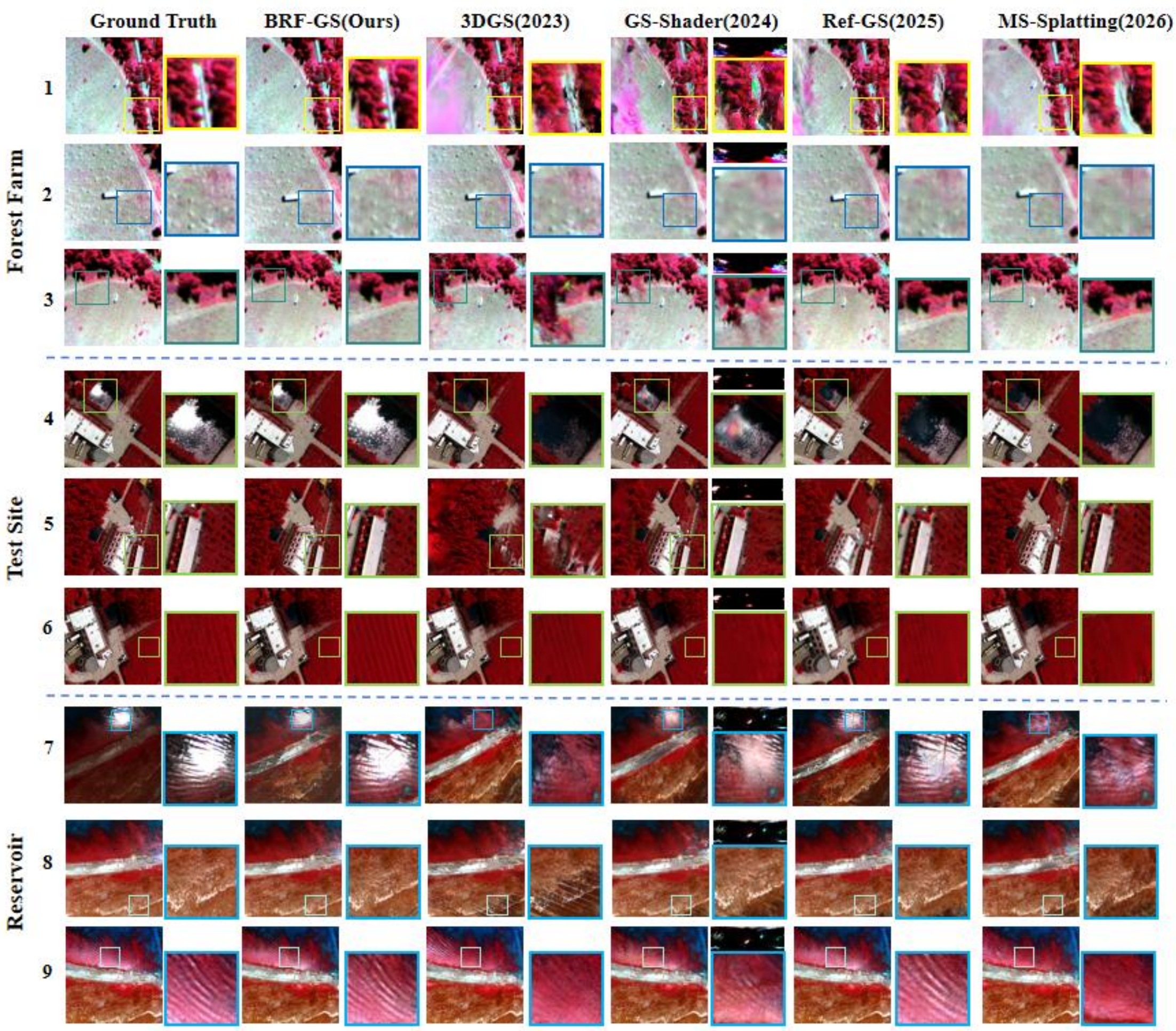

**Fig**. 6 Comparison of BRF image generation with other state-of-the-art methods. For detailed comparison, zoomed-in views are provided for the vegetation-covered road, haystack, and vegetation–grassland junction in the forest farm scene; the pond surface, buildings, and field ridges in the test site scene; and the specular reflection region on the water surface, texture-rich tidal flats, and water ripples in the reservoir scene. As GaussianShader decouples ambient illumination from material properties, the corresponding decoupled ambient illumination maps are shown in the upper-right corners of the GaussianShader results.

Fig. 6 presents qualitative comparisons of novel-view BRF image generation across the three scenes. For each scene, representative regions with different directional reflectance and structural characteristics are enlarged, including vegetation-covered roads, haystacks, and vegetation–grassland boundaries in the Forest Farm; pond surfaces, buildings, and field ridges in the Test Site; and specular reflection regions, texture-rich tidal flats, and water ripples in the Reservoir. For GaussianShader, the estimated ambient illumination maps are also shown in the upper-right corners of the corresponding results. Overall, the comparison methods exhibit different limitations in representing directional reflectance and scene structure.

The original 3DGS has limited capacity to represent high-frequency view-dependent reflectance due to its low-order spherical harmonics (SH) formulation. As a result, it cannot adequately reproduce the pronounced water glints over the pond in the test site scene (enlarged view in row 4, column 3) or the specular glints in the reservoir scene (row 7, column 3). Noticeable artifacts are also observed over vegetation, where complex canopy structure and view-dependent reflectance lead to substantial radiometric differences and ambiguity across viewing angles (enlarged views in rows 1, 3, and 5, column 3).

GaussianShader adopts a microfacet BRDF model and therefore provides improved modeling of specular appearance, allowing it to reproduce water glint more effectively. However, its specular-oriented reflectance model is less suitable for vegetation, whose directional reflectance is strongly influenced by canopy geometry and shadowing. This results in noticeable artifacts in vegetation regions (rows 1, 3, and 5, column 4). The ambient illumination estimated (upper-right corners) differs substantially from the illumination expected under natural outdoor conditions, which is governed by direct solar radiation and diffuse atmospheric illumination. This discrepancy suggests a fundamental limitation of modeling the entire scene with a single microfacet BRDF: distinct surface types and reflection mechanisms are forced into the same reflectance model, resulting in a physically inconsistent estimation of ambient illumination.

Ref-GS employs a more accurate normal representation based on 2DGS and deferred shading to improve geometric and directional reflectance modeling. It produces clearer geometric structures and finer textures in features such as roads, field ridges, and water ripples, while also reproducing water glints more effectively (enlarged view in row 7, column 5). However, 2D Gaussian primitives as a planar representation are less suitable for scenes dominated by complex 3D structures. In particular, the vegetation canopy may not be adequately represented, resulting in blurred structures and artifacts in vegetation regions (row 1, column 5).

MS-Splatting extends Gaussian splatting to spectral imagery, but it does not explicitly model directional reflectance. As a result, it produces noticeable artifacts in vegetation regions and has limited ability to reproduce high-frequency water reflections observed in the test site and reservoir scenes.

In contrast, BRF-GS provides more adaptive and effective modeling of directional reflectance across different land-cover types. It effectively captures the strong specular response of water surfaces and the complex directional reflectance of vegetation resulting from the combined effects of canopy structure and volumetric scattering, substantially reducing the artifacts observed with the comparison methods. In terms of geometric structure and texture, BRF-GS preserves fine details in both relatively planar surfaces, such as field ridges, roads, and water ripples, and structurally complex vegetation canopies.

#### *4.2.2 Quantitative Comparison*

The qualitative comparisons show that BRF-GS provides more consistent reconstruction of directional reflectance, geometric structures, and fine-scale textures across diverse land-cover types. To further quantify these differences, Table 1 reports the novel-view image generation and spectral reconstruction performance across the three scenes using PSNR, SSIM, LPIPS, and spectral angle mapper (SAM). PSNR, SSIM, and LPIPS were calculated from the CIR composite images, whereas SAM was calculated over the spectral bands used for training. Because the RGB-image-based methods do not reconstruct spectral reflectance, SAM is not applicable to these methods and is denoted by "–".

Table 1 Quantitative comparison of novel view BRF image generation performance. The best results are shown in bold and the second-best results are underlined.

| **Scene** | **Forest Farm** | | | | **Test Site** | | | | **Reservoir** | | | |
|---|---|---|---|---|---|---|---|---|---|---|---|---|
| **Metrics** | **PSNR↑** | **SSIM↑** | **LPIPS↓** | **SAM↓** | **PSNR↑** | **SSIM↑** | **LPIPS↓** | **SAM↓** | **PSNR↑** | **SSIM↑** | **LPIPS↓** | **SAM↓** |
| 3DGS | 20.12 | 0.556 | 0.474 | - | 19.83 | 0.538 | 0.489 | - | 19.57 | 0.520 | 0.493 | - |
| GaussianShader | 20.55 | 0.577 | 0.451 | - | 21.31 | 0.601 | 0.423 | - | 21.32 | 0.619 | 0.400 | - |
| Ref-GS | <u>22.65</u> | <u>0.701</u> | <u>0.294</u> | - | <u>23.32</u> | <u>0.729</u> | <u>0.267</u> | - | <u>23.87</u> | <u>0.744</u> | <u>0.256</u> | - |
| MS-Splatting | 21.14 | 0.634 | 0.406 | <u>0.278</u> | 20.78 | 0.618 | 0.412 | <u>0.291</u> | 20.45 | 0.601 | 0.425 | <u>0.304</u> |
| BRF-GS (Ours) | **24.54** | **0.734** | **0.232** | **0.112** | **24.63** | **0.772** | **0.241** | **0.119** | **24.16** | **0.758** | **0.255** | **0.128** |

As shown in Table 1, BRF-GS consistently achieves the best performance across all three scenes and all evaluated metrics, demonstrating its effectiveness for BRF image generation. These improvements indicate that BRF-GS provides substantially better reconstruction fidelity and perceptual consistency across scenes with different land-cover compositions.

More importantly, BRF-GS exhibits a clear advantage in spectral fidelity. Among the compared methods, MS-Splatting is the only baseline that explicitly reports SAM, whereas the other methods do not provide physically meaningful spectral reflectance outputs for this evaluation. BRF-GS achieves SAM values of 0.112, 0.119, and 0.128 across the three scenes, indicating that the reconstructed spectra remain closely aligned with the reference reflectance in spectral shape. Compared with MS-Splatting, which obtains SAM values of 0.278, 0.291, and 0.304, BRF-GS reduces the spectral angular error by approximately 59.7%, 59.1%, and 57.9%, respectively. This substantial improvement suggests that explicitly modeling directional reflectance, rather than relying on conventional view-dependent color representations, is important for preserving the spectral characteristics of hyperspectral observations under varying viewing directions.

The PSNR gain reaches 4.80 dB in the Test Site scene, compared with 4.42 dB and 4.59 dB in the Forest Farm and Reservoir scenes, respectively. The Forest Farm scene is primarily composed of vegetation, while the Reservoir scene is dominated by water surfaces, and thus their reflectance characteristics are relatively concentrated around specific types of directional responses. In contrast, the Test Site contains a heterogeneous mixture of natural and artificial surfaces, including vegetation, ground, buildings, and metallic objects, with pronounced 3D structures and substantially different reflection mechanisms. The larger improvement of BRF-GS in this scene indicates that its advantage becomes more pronounced as the diversity and spatial complexity of directional reflectance increase. Rather than being specialized for a particular surface type, the proposed hybrid-driven representation provides a more robust mechanism for jointly modeling heterogeneous and coexisting reflection characteristics in complex remote sensing scenes.

### *4.3. Spectral Reconstruction Accuracy Analysis of Representative Surface Types*

To further evaluate the spectral reflectance reconstruction accuracy of BRF-GS across different land-cover and surface types, we selected six representative targets from the test images: water, bare soil (including soil and sand), vehicle, vegetation (including shrubs and coniferous forest), building, and dry grass. For each target, a region of interest (ROI) was manually delineated, and the spectral reflectance values of all pixels within the ROI were extracted. Pixel-wise outlier removal was then performed before calculating the mean spectrum. Fig. 7 presents the mean reconstructed and ground truth spectral curves, together with their logarithmic errors, for the six representative targets across all testing angles. Logarithmic error is used instead of conventional relative error to better visualize the error magnitude while avoiding excessive amplification in regions with extremely low reflectance.

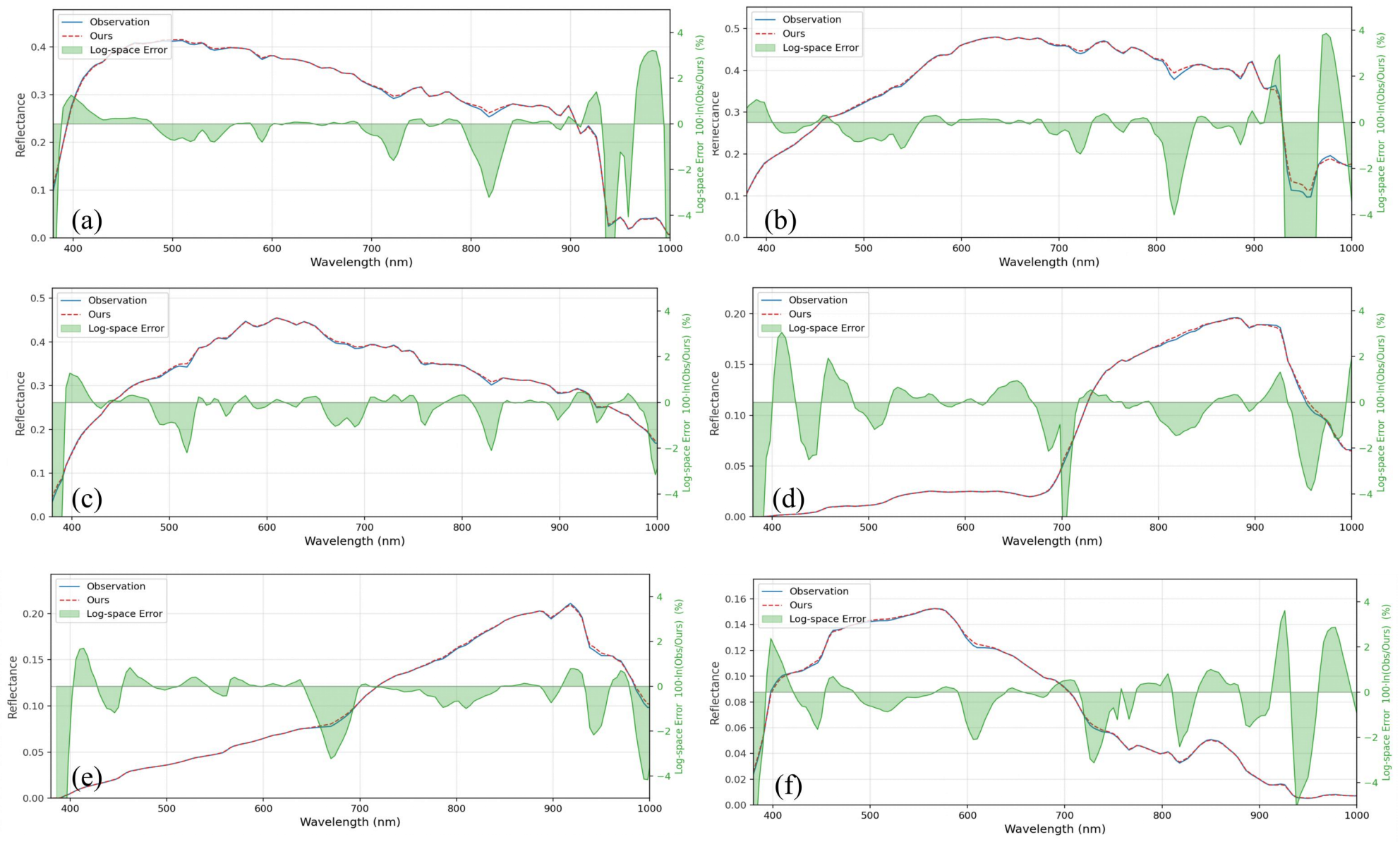

Fig. 7 Mean reconstructed and ground truth spectral reflectance curves, together with their logarithmic errors. The vertical scales of the spectral reflectance plots differ among the subplots because of the substantially different reflectance magnitudes of the surface types. (a) Vehicle; (b) Bare soil; (c) Building; (d) Vegetation; (e) Dry grass; (f) Water body.

The reconstructed spectral curves show high consistency with the ground truth spectra in both overall spectral trends and characteristic absorption features. As reflectance approaches zero, logarithmic error becomes sensitive to small absolute deviations because of the small denominator, resulting in artificially amplified error values. Excluding these low-reflectance regions, the remaining deviations are mainly concentrated around rapidly varying spectral features, particularly absorption features, where the reconstructed spectra are slightly smoother than the ground truth. This smoothing may be attributed to the loss of fine-scale spectral variations during latent-space compression. A relatively larger error is also observed near strong atmospheric absorption bands, particularly around the approximately 940-nm water-vapor absorption feature. This effect is mainly associated with the radiometric calibration procedure. The white-panel calibration converts the measured sensor responses into apparent reflectance but does not explicitly compensate for atmospheric absorption along the observation path. Because the observation geometry varies among acquisition angles, the corresponding atmospheric optical paths also vary, which can introduce wavelength-dependent discrepancies near strong atmospheric absorption features. The error in this region is generally larger for the objects (water and bare soil) in the reservoir scene than for the objects (dry grass and buildings) in the forest farm scene, likely reflecting differences in atmospheric conditions and optical path lengths between the two scenes.

Reconstruction accuracy also varies across surface types. Excluding extremely low-reflectance regions and atmospheric absorption bands, vegetation and water generally exhibit larger reconstruction errors. Vegetation canopies have more complex vertical distributions and fine structures than relatively simple surface targets, which increases the difficulty of accurately reconstructing their 3D structures. For water, the observed signal is affected not only by surface reflection but also by transmission and volumetric light transport within the water column, which are not explicitly modeled by the current directional-reflectance kernel. These unmodeled effects may therefore be partially absorbed into the estimated reflection components, resulting in additional reconstruction errors. Compared with surface and subsurface scattering, modeling volumetric light transport requires accounting for light propagation and interactions along the observation path, which is beyond the scope of the current surface-reflectance formulation.

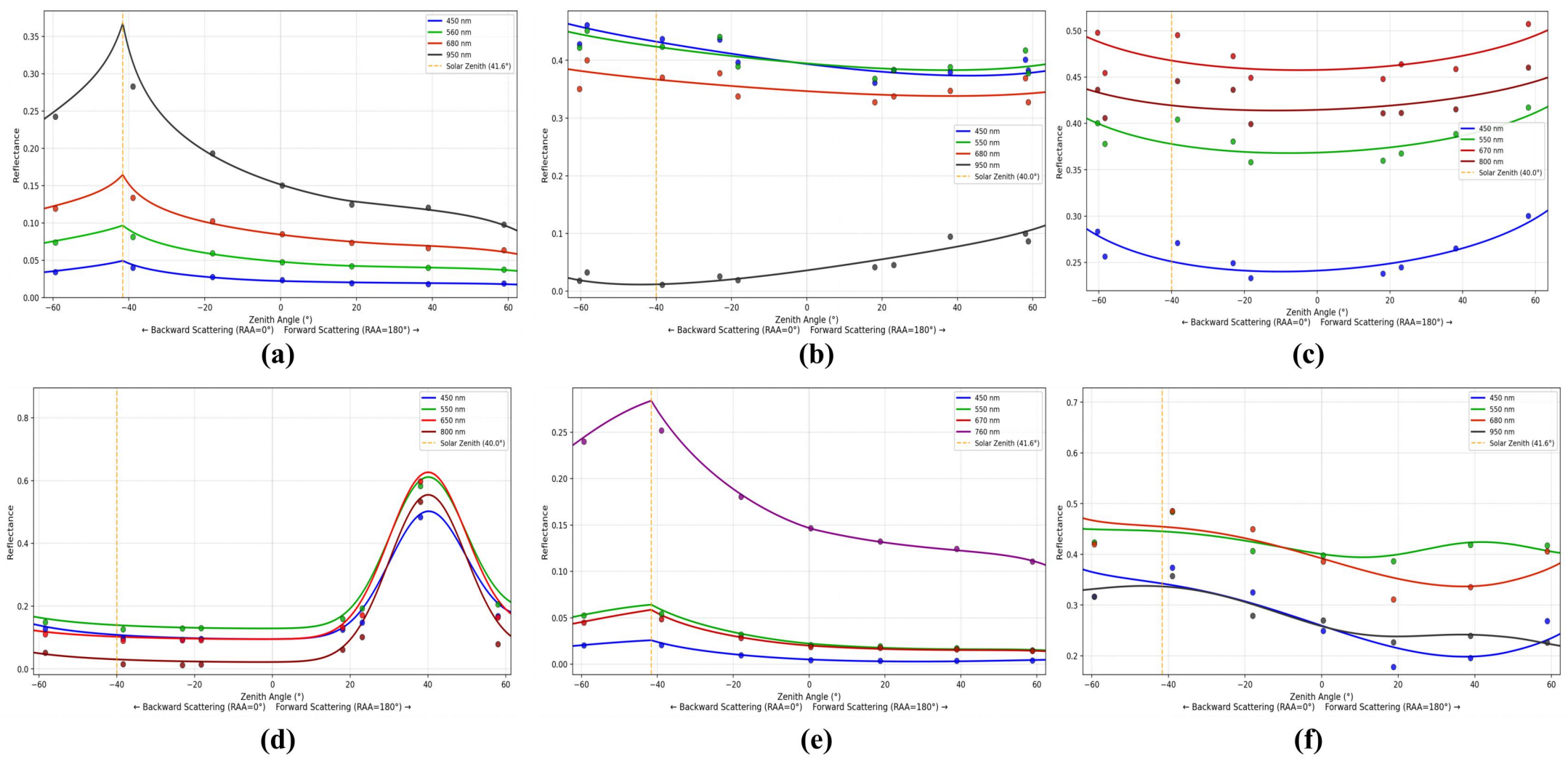

Fig. 8 Fitted spectral curves and measured values at representative bands for six representative surface types along the principal plane. The solar zenith is indicated by an orange dashed line. (a) Dry grass; (b) Building; (c) Bare soil; (d) Water body; (e) Forest; (f) Vehicle.

The principal plane defined by the solar illumination direction generally exhibits the strongest anisotropic and directional variations in surface reflectance (Sandmeier et al., 1998). We therefore further evaluated the ground-truth and reconstructed reflectance at different zenith angles along the principal plane, with the results for the six representative surface types shown in Fig. 8. Among natural land-covers, dry grass and forest exhibit pronounced hotspot and bowl effects (Kuusk, 1991), while water exhibits a pronounced peak in the specular-reflection direction, which are all well reproduced by the proposed model. For artificial target surfaces, the model also closely matches the observed directional reflectance, demonstrating its ability to represent diverse and complex bidirectional reflection characteristics.

Due to practical constraints during data acquisition, exact alignment between the sampling points and the theoretical principal plane could not always be achieved, as the solar position varied over the course of the experiment while the flight path was predetermined. For visualization, the observation closest to the theoretical principal-plane position was selected as the visualized points in Fig. 8. This angular mismatch is the primary source of the slight discrepancies between some measured points and the fitted curves in Fig. 8.

Overall, BRF-GS accurately reconstructs the directional reflectance characteristics of diverse surface types, including their spectral features and anisotropic responses along the principal plane.

### *4.4. Analysis of Directional Reflectance Distribution*

This section evaluates the capability of BRF-GS to reconstruct the directional reflectance distributions of different surface types.

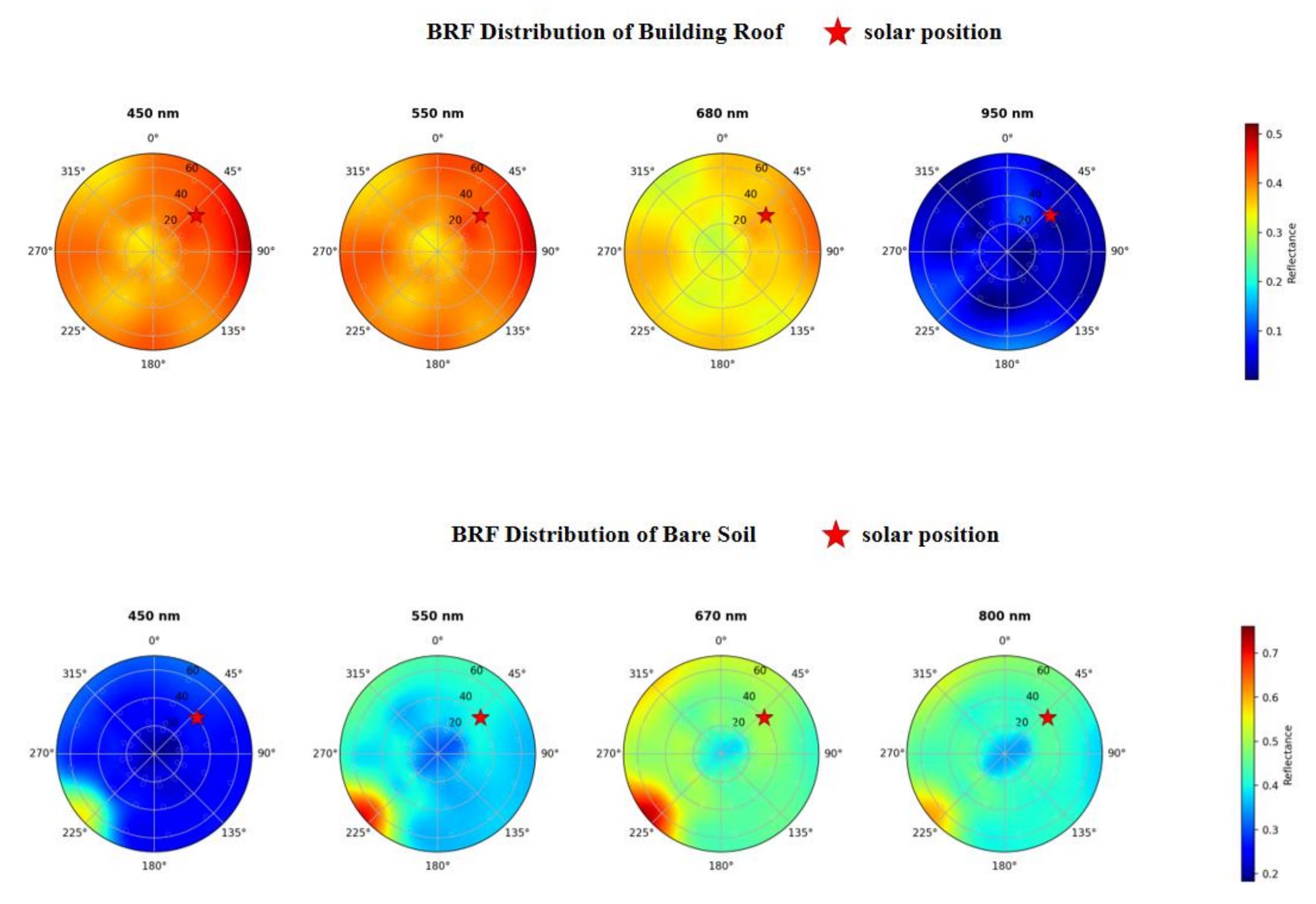

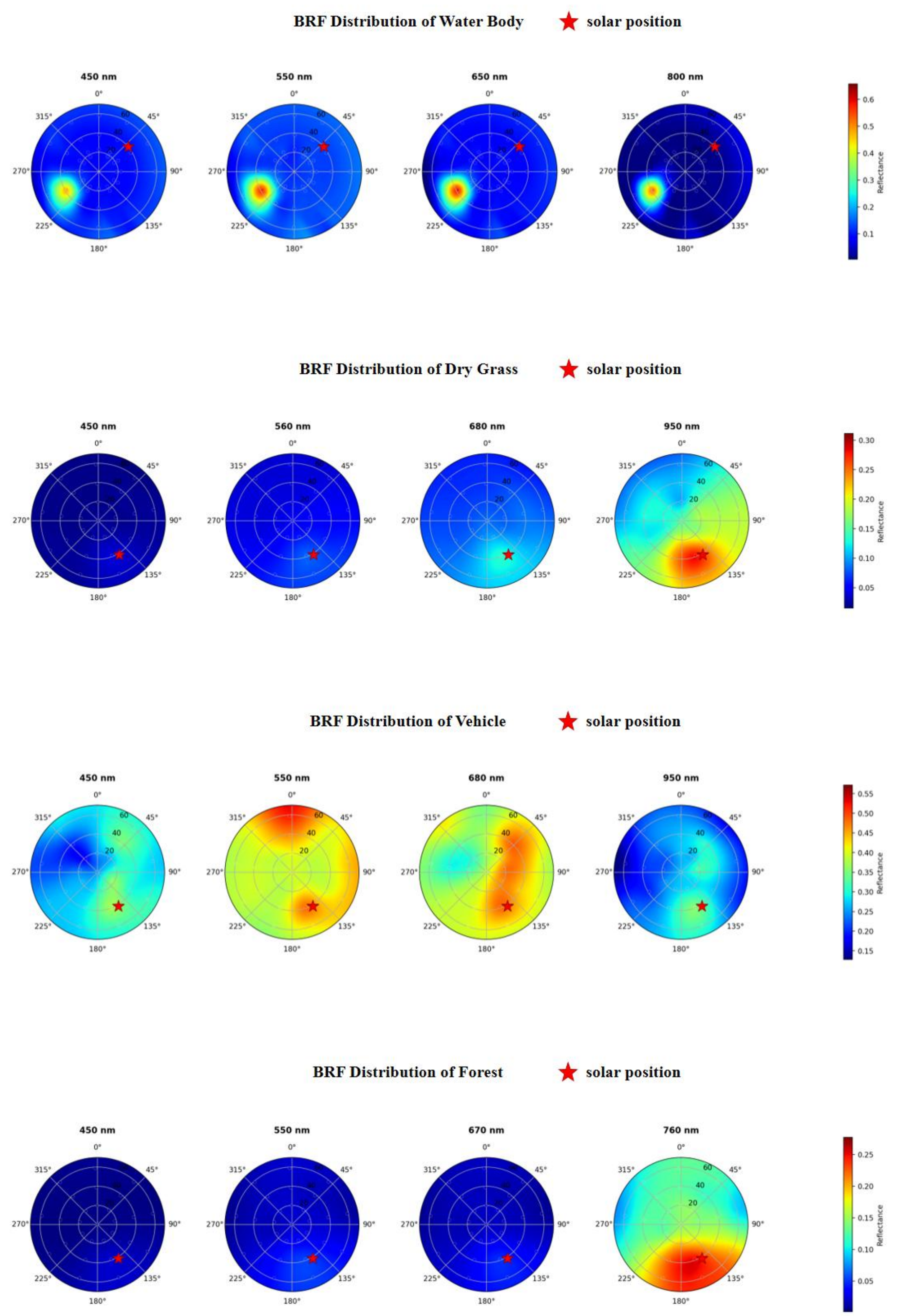

Fig. 9 Heatmaps of the directional reflectance distribution at representative spectral bands for the six surface types. The solar direction is indicated by a red star.

Fig. 9 presents polar heatmaps of the reconstructed directional reflectance distributions at representative spectral bands for the six surface types. The results show that BRF-GS reconstructs continuous directional reflectance distribution beyond the discrete angular observations from the measurements. The reconstructed distributions also reveal complex anisotropic responses that cannot be adequately captured by simple parametric BRDF models. Distinct directional patterns are observed across different surface types, reflecting their underlying physical and structural properties.

Water exhibits a pronounced specular peak in the specular-reflection direction across the analyzed bands, with reflectance strongly concentrated around this direction. Forest exhibits pronounced backscattering enhancement and a more isotropic response at 760 nm, consistent with the strong multiple-scattering effects in the near-infrared region (Ross, 1981). In contrast, their visible-band responses are relatively weak and exhibit less pronounced directional variation. The hotspot of coniferous forest shows a spatially extended distribution rather than an idealized point-like peak, which is consistent with the multi-scale clumping structure of coniferous canopies (Chen et al., 2005). This structure broadens the angular range over which the hotspot response is observed.

Dry grass exhibits a more pronounced hotspot at 680 nm, where chlorophyll absorption is weaker. In the near-infrared region, its hotspot becomes more concentrated, indicating stronger angular sensitivity than coniferous forest.

Soil exhibits a pronounced backscattering enhancement across the analyzed spectral bands, with a broad reflectance peak centered approximately on the solar-backscattering direction. This hotspot behavior is associated with the reduced visibility of surface shadows when the viewing direction approaches the backscattering direction, and is consistent with the well-documented non-Lambertian

BRDF characteristics of bare soil surfaces (Hapke et al., 1996). The relative minimum near nadir further indicates the strong angular anisotropy of the soil surface. The backscattering enhancement becomes more pronounced in the near-infrared bands, consistent with the stronger directional contrast commonly observed in soil reflectance at longer wavelengths (Pinty et al., 1989).

Artificial targets, including building roofs and vehicle bodies, exhibit asymmetric and multimodal directional distributions due to their complex geometric structures and material compositions. Building roofs, which are primarily composed of concrete and cement, show relatively weak directional variation, whereas vehicles, which are primarily composed of metal, exhibit a more pronounced specular component.

### *4.5. Ablation Studies*

To evaluate the effectiveness of the major components of BRF-GS, we conduct ablation studies on three key design choices: geometry-reliable spectral band selection, the hybrid BRDF-driven directional reflectance kernel, and the geometry–reflection decoupled training strategy.

#### *Ablation of Geometry-reliable Band Selection*

To evaluate the effectiveness of the proposed geometry-reliable spectral band selection strategy, we conduct ablation experiments on the Forest Farm scene. Specifically, we compare eight band selection strategies for geometric initialization: All bands, RGB, CIR, Central band, Max-SNR band, Max-feature-point band, Top-1 geometry-reliable band, and Top-3 geometry-reliable bands (Ours). For the All-band setting, local features are independently extracted from all available spectral bands and then aggregated across bands with duplicate feature points removed. For the RGB and CIR settings, three representative bands are selected based on the center wavelengths of the corresponding spectral ranges. Specifically, RGB uses bands centered at 450, 550, and 650 nm, while CIR uses 550, 650, and 850 nm. The Central-band setting uses a single band at 675 nm. The Max-SNR and Max-feature-point settings select the bands with the highest SNR and the largest number of detected feature points, respectively, providing single-factor baselines for evaluating the proposed geometry-reliable criterion. Top-1 geometry-reliable band selects the highest-ranked band, whereas Top-3 geometry-reliable bands use the three highest-ranked bands.

We evaluate the geometric reliability of different spectral configurations using the average number of detected feature points and the feature matching success rate across multi-views. These metrics characterize the availability and consistency of local geometric features before SfM reconstruction. We further report the resulting BRF image quality in terms of PSNR, SSIM, and SAM to examine whether improved geometric reliability translates into better hyperspectral BRF reconstruction. All subsequent Gaussian optimization and rendering settings are kept identical across different configurations.

Table 2 Ablation study of geometry-reliable band selection strategy (Under forest farm scene)

| **Model setting** | PSNR↑ | SSIM↑ | SAM↓ | Detected feature points ↑ | Matching success rate ↑ |
|---|---|---|---|---|---|
| All bands (350 – 1000 nm) | 23.21 | 0.685 | 0.125 | **10642** | 82.7% |
| RGB bands (650 nm, 550 nm, 450 nm) | 23.61 | 0.702 | 0.123 | 3621 | 96.8% |
| CIR bands (850 nm, 650 nm, 550 nm) | 24.25 | 0.724 | 0.117 | 4613 | 97.6% |
| Central band (675 nm) | 21.48 | 0.621 | 0.189 | 822 | 97.3% |
| Max SNR band (870 MHz) | 24.27 | 0.723 | 0.116 | 2238 | **98.7%** |
| Max feature points band (950 nm) | 23.14 | 0.683 | 0.128 | 9033 | 88.0% |
| 1 Geometry-reliable band | 24.49 | 0.730 | 0.113 | 8525 | 98.2% |
| 3 Geometry-reliable bands (Ours) | **24.54** | **0.734** | **0.112** | 9227 | 98.4% |

As shown in Table 2, the results reveal that the number of detected feature points or spectral bands is not a direct indicator of geometric reliability. Increasing the number of input bands does not necessarily improve reconstruction, as low-information or noisy bands may introduce unreliable features, dilute effective geometric information in multi-band representations, and increase the difficulty of cross-view correspondence establishment. For spectrally informative bands, a single band already provides a comparable number of effective features to a three-band combination, suggesting limited benefit from further increasing the number of bands. Moreover, the band with the highest SNR does not necessarily yield the most feature points or the best reconstruction performance, indicating that SNR alone is insufficient to characterize the geometric utility of a spectral band. Conversely, bands with insufficient spatial information, such as the central band (675nm) in our experiment, produce substantially fewer features and lead to degraded reconstruction. Interestingly, selecting the band with the largest number of detected features also does not consistently yield the best performance, further demonstrating that feature quantity alone cannot adequately characterize the quality of geometric cues. Overall, these results support the necessity of selecting spectral bands based on their geometric reliability rather than simply maximizing the number of bands, SNR, or detected feature points.

***Ablation of Directional Reflectance Kernels***

To evaluate the effectiveness of the proposed hybrid BRDF-driven directional reflectance kernel, we conduct ablation experiments by systematically removing individual kernel components and comparing them with the conventional spherical harmonics (SH) representation. We study five settings: w/o Isotropic, w/o Volumetric, w/o Geometric, w/o Specular, and the full BRF-GS model containing all four components. In addition, a conventional SH representation is used as a baseline, where the directional reflectance is modeled using spherical harmonics under the same Gaussian representation and training settings.

For each ablation variant, only the directional reflectance representation is modified, while the geometry initialization, two-stage training strategy, Gaussian representation, optimization settings, and spectral modeling components are kept identical. The four component-wise ablations evaluate the contribution of each physically motivated kernel by removing the corresponding component from the full model.

We evaluate all variants using PSNR and SAM for hyperspectral BRF image generation. These experiments aim to determine whether the proposed BRDF-driven kernel provides a more effective representation of complex directional reflectance than conventional SH and whether the four physically motivated components make complementary contributions to hyperspectral BRF reconstruction.

Table 3 Ablation study of directional reflectance kernels

| **Model setting** | PSNR↑ | SAM↓ |
|---|---|---|
| Baseline (spherical harmonics) | 21.56 | 0.234 |
| Ours w/o Iso | 20.48 | 0.277 |
| Ours w/o Vol | 22.72 | 0.182 |
| Ours w/o Geo | 22.84 | 0.180 |
| Ours w/o Spec | 23.76 | 0.138 |
| BRF-GS (Full) | **24.44** | **0.120** |

As shown in Table 3, the results demonstrate that all four kernels contribute complementarily to directional reflectance modeling. Removing any individual kernel consistently degrades the reconstruction performance, while the full BRF-GS model achieves the highest PSNR and lowest SAM. In particular, removing the isotropic component causes the largest performance drop, highlighting its fundamental role in modeling the overall reflectance response. The volumetric scattering and geometric-optical components also provide substantial and comparable improvements, reflecting the importance of volume- and geometry-dependent directional effects in remote sensing scenery. In contrast, removing the specular component results in a relatively smaller degradation. Moreover, BRF-GS substantially outperforms the spherical-harmonics baseline, demonstrating the advantage of leveraging BRDF kernels derived from long-established empirical and semi-empirical knowledge of land-surface directional reflectance, rather than relying on generic basis functions.

***Ablation of Geometry–Reflectance Decoupled Training***

To investigate whether decoupling geometry and radiation optimization can alleviate the optimization ambiguity between geometric structure and directional reflectance, we compare the proposed two-stage training strategy with joint optimization under otherwise identical settings.

Table 4 Ablation study of geometry–reflectance decoupled training

| **Model setting** | PSNR↑ | SAM↓ |
|---|---|---|
| Joint optimization | 24.07 | 0.188 |
| BRF-GS (Full) | **24.44** | **0.120** |

As shown in Table 4, the comparison between joint and decoupled optimization demonstrates the benefit of separating geometry and radiation optimization. Compared with joint optimization, BRF-GS achieves a modest improvement in PSNR from 24.07 to 24.44 dB, but a substantially larger reduction in SAM from 0.188 to 0.120. This indicates that the decoupled training strategy particularly improves spectral fidelity rather than merely reducing image-level reconstruction error. Jointly optimizing geometry and directional reflectance may introduce optimization ambiguity, allowing geometric and radiometric parameters to compensate for each other. By first establishing the scene geometry and subsequently optimizing the full spectral directional reflectance with the geometry fixed, BRF-GS reduces such coupling and provides a more stable basis for accurate hyperspectral BRF reconstruction.

### *4.6. Comparison with 3D radiative transfer simulation methods*

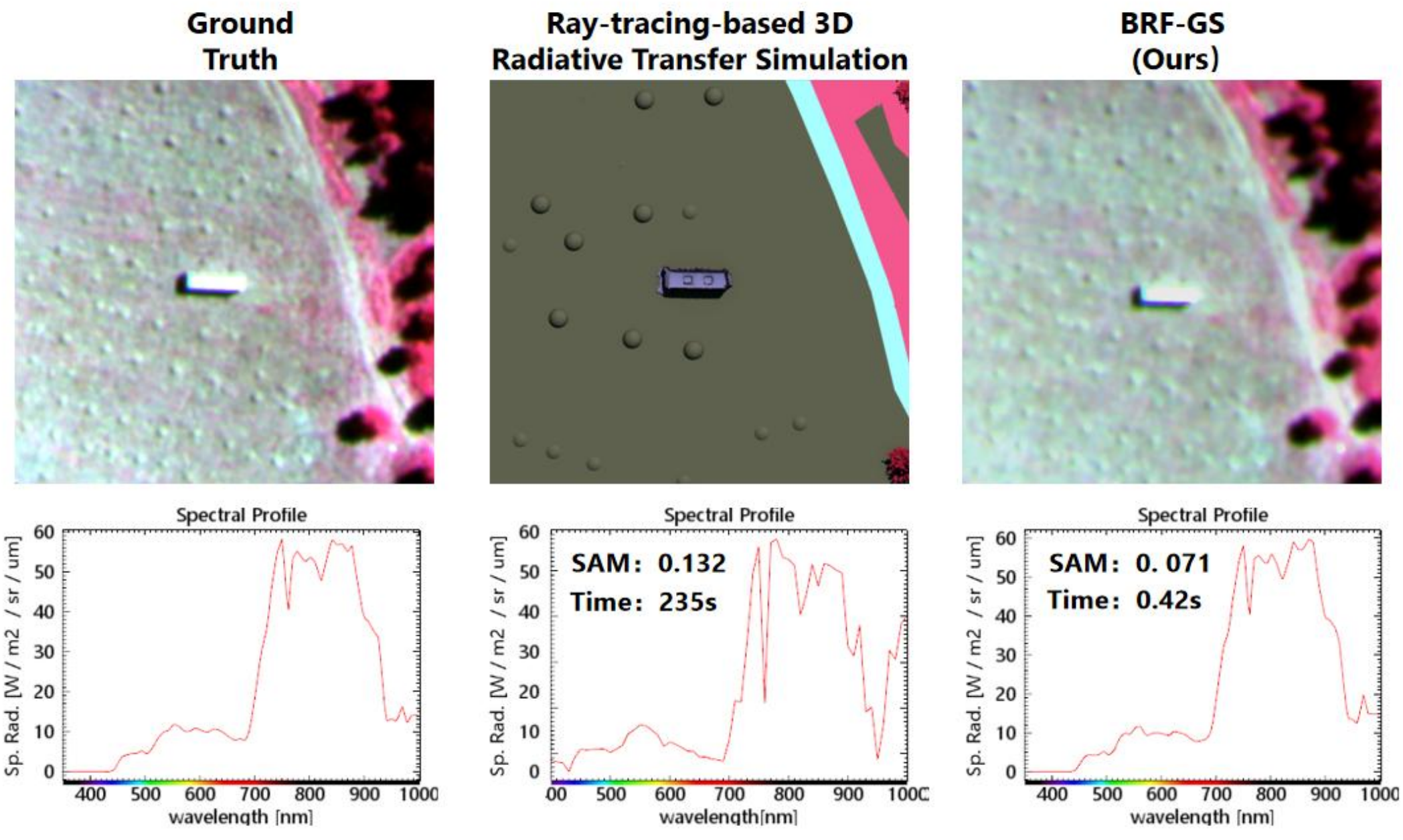

Fig. 10 Comparisons with a 3D radiative transfer simulation method in both BRF image rendering and spectral reconstruction.

Finally, to further assess the accuracy and efficiency of our method against the 3D radiative transfer simulation method, we compared it with a hyperspectral radiative transfer simulation method based on ray tracing, using a manually constructed 3D radiative transfer model (Li et al., 2025) of the forest farm scene. The model contains 1.4 million facets and divides the scene into 12 material classes, with material parameters assigned to each class. Ground objects belonging to the same material class are assumed to share identical material parameters. The output image resolution is 350 × 350 pixels, with 2,000 rays traced per pixel and a maximum of four reflection bounces per ray.

As shown in Fig. 10, our method produces hyperspectral BRF images and spectral curves that are closer to the ground truth. In terms of computational efficiency, with GPU acceleration, the ray-tracing-based method requires 235 s (RTX 4080) to generate a single 164-band BRF image at the same viewpoint, whereas our method renders the corresponding frame in only 42 ms (RTX A6000) .

## 5. CONCLUSIONS AND FUTURE WORK

### *5.1.Conclusions*

This study addressed the critical challenges in remote sensing hyperspectral BRF modeling and image generation, specifically the complex directional reflectance of heterogeneous ground objects, high computational costs, and data scarcity. To overcome these hurdles, we proposed BRF-GS, a novel framework for hyperspectral bidirectional reflectance factor reconstruction and image generation based on 3D Gaussian splatting, and established the AIR-BRF dataset to support multi-angle hyperspectral analysis.

Our experiments demonstrate that BRF-GS significantly outperforms existing state-of-the-art methods in both image generation visual quality and spectral fidelity. Notably, the proposed hybrid BRDF-driven directional reflectance kernel and the geometry–reflectance decoupled two-stage training strategy proved effective in minimizing spectral distortion, as evidenced by superior performance in PSNR, SSIM, and specifically the SAM metric. In terms of directional reflectance patterns, BRF-GS effectively captures the distinct angular responses of different ground objects—such as the vegetation hotspot effect and water surface specular reflection. These results demonstrate that the proposed hybrid-driven directional reflectance representation can effectively characterize the diverse BRF variation patterns of heterogeneous ground objects within a scene-level 3D representation.

Overall, this study demonstrates the potential of 3D Gaussian Splatting for scene-level hyperspectral directional reflectance modeling and efficient BRF image generation in high-spatial-resolution remote sensing. The AIR-BRF dataset provides a data foundation for hyperspectral directional reflectance modeling, algorithm evaluation, and multi-angle remote sensing data generation. Meanwhile, BRF-GS provides a new approach for constructing scene-level 3D directional reflectance representations from measured multi-angle hyperspectral observations, offering a foundation for subsequent quantitative remote sensing analysis, directional reflectance characterization, and related biophysical and environmental parameter retrieval.

### *5.2. Future Work*

Although BRF-GS has achieved good results in hyperspectral BRF image generation and spectral reconstruction, the following directions are still worth further exploration:

**Atmospheric-coupled radiative transfer modeling.** This paper primarily models apparent reflectance obtained after ground-based radiometric calibration without explicit correction for atmospheric absorption and scattering along different atmospheric propagation paths under different viewing geometries. In future work, atmospheric and surface radiative transfer could be jointly modeled to account for the effects of varying observation paths and atmospheric conditions, thereby establishing a more physically consistent link between remote sensing observations and surface directional reflectance.

**Extended validation across diverse land-cover types.** Owing to limitations in the spatial and seasonal coverage of the dataset, this study has not yet included land-cover types with distinctive directional reflectance characteristics, such as deserts and snow/ice surfaces. In the future, the multi-angle hyperspectral BRF dataset could be expanded to encompass a broader range of regions, seasons, and land-cover types, enabling a more systematic evaluation of the model's generalization capability and robustness across diverse surface conditions.

## REFERENCE

[1] Barnsley, M.J., Settle, J.J., Cutter, M.A., Lobb, D.R., Teston, F., 2004. The PROBA/CHRIS mission: a low-cost smallsat for hyperspectral multiangle observations of the Earth surface and atmosphere. IEEE Trans. Geosci. Remote Sensing 42, 1512–1520. https://doi.org/10.1109/TGRS.2004.827260

[2] Zhang, B., Liu, Q.H., Li, X., Liu, L., Yang, B., Husi, L., Gao, L., Zhang, W., Zhang, H., Bian, Z., Qi, M., Chen, C., Shang, H., 2025. The core concepts and fundamental issues of remote sensing science. National Remote Sensing Bulletin 29, 1–48. https://doi.org/10.11834/jrs.20244503

[3] Burkart, A., Aasen, H., Alonso, L., Menz, G., Bareth, G., Rascher, U., 2015. Angular Dependency of Hyperspectral Measurements over Wheat Characterized by a Novel UAV Based Goniometer. Remote Sensing 7, 725–746. https://doi.org/10.3390/rs70100725

[4] Chen, J.M., Leblanc, S.G., 1997. A four-scale bidirectional reflectance model based on canopy architecture. IEEE Trans. Geosci. Remote Sensing 35, 1316–1337. https://doi.org/10.1109/36.628798

[5] Chen, J.M., Menges, C.H., Leblanc, S.G., 2005. Global mapping of foliage clumping index using multi-angle satellite data. Remote Sensing of Environment 97, 447–457. https://doi.org/10.1016/j.rse.2005.05.003

[6] Cook, R.L., Torrance, K.E., 1982. A Reflectance Model for Computer Graphics. ACM Trans. Graph. 1, 7–24. https://doi.org/10.1145/357290.357293

[7] Deng, W., Wu, T., Ni, Z., Liu, Y., Jia, H., Ling, Q., 2026. Hyperspectral imaging

meets 3D Gaussian Splatting: A novel approach beyond 3D plant morphology. Plant Phenomics 8, 100198. https://doi.org/10.1016/j.plaphe.2026.100198

[8] Deschamps, P.-Y., Breon, F.-M., Leroy, M., Podaire, A., Bricaud, A., Buriez, J.-C., Seze, G., 1994. The POLDER mission: instrument characteristics and scientific objectives. IEEE Trans. Geosci. Remote Sensing 32, 598–615. https://doi.org/10.1109/36.297978

[9] Gastellu-Etchegorry, J.-P., Lauret, N., Yin, T., Landier, L., Kallel, A., Malenovsky, Z., Bitar, A.A., Aval, J., Benhmida, S., Qi, J., Medjdoub, G., Guilleux, J., Chavanon, E., Cook, B., Morton, D., Chrysoulakis, N., Mitraka, Z., 2017. DART: Recent Advances in Remote Sensing Data Modeling With Atmosphere, Polarization, and Chlorophyll Fluorescence. IEEE J. Sel. Top. Appl. Earth Observations Remote Sensing 10, 2640–2649. https://doi.org/10.1109/JSTARS.2017.2685528

[10] Goodenough, A.A., Brown, S.D., 2017. DIRSIG5: Next-Generation Remote Sensing Data and Image Simulation Framework. IEEE J. Sel. Top. Appl. Earth Observations Remote Sensing 10, 4818–4833. https://doi.org/10.1109/JSTARS.2017.2758964

[11] Hao, D., Wen, J., Xiao, Q., You, D., Tang, Y., 2020. An Improved Topography-Coupled Kernel-Driven Model for Land Surface Anisotropic Reflectance. IEEE Trans. Geosci. Remote Sensing 58, 2833–2847. https://doi.org/10.1109/TGRS.2019.2956705

[12] Hapke, B., DiMucci, D., Nelson, R., Smythe, W., 1996. The cause of the hot spot in vegetation canopies and soils: Shadow-hiding versus coherent backscatter. Remote Sensing of Environment 58, 63–68. https://doi.org/10.1016/0034-

4257(95)00257-X

[13] Huang, B., Yu, Z., Chen, A., Geiger, A., Gao, S., 2024. 2D Gaussian Splatting for Geometrically Accurate Radiance Fields, in: Special Interest Group on Computer Graphics and Interactive Techniques Conference Conference Papers. Presented at the SIGGRAPH '24: Special Interest Group on Computer Graphics and Interactive Techniques Conference, ACM, Denver CO USA, pp. 1–11. https://doi.org/10.1145/3641519.3657428

[14] Ientilucci, E.J., Brown, S.D., 2003. Advances in wide-area hyperspectral image simulation, in: Watkins, W.R., Clement, D., Reynolds, W.R. (Eds.), . Presented at the AeroSense 2003, Orlando, FL, p. 110. https://doi.org/10.1117/12.488706

[15] Jiang, Y., Tu, J., Liu, Y., Gao, X., Long, X., Wang, W., Ma, Y., 2024. GaussianShader: 3D Gaussian Splatting with Shading Functions for Reflective Surfaces, in: 2024 IEEE/CVF Conference on Computer Vision and Pattern Recognition (CVPR). Presented at the 2024 IEEE/CVF Conference on Computer Vision and Pattern Recognition (CVPR), IEEE, Seattle, WA, USA, pp. 5322–5332. https://doi.org/10.1109/CVPR52733.2024.00509

[16] Kerbl, B., Kopanas, G., Leimkuehler, T., Drettakis, G., 2023. 3D Gaussian Splatting for Real-Time Radiance Field Rendering. ACM Trans. Graph. 42, 1–14. https://doi.org/10.1145/3592433

[17] Kuusk, A., 1991. The Hot Spot Effect in Plant Canopy Reflectance, in: Myneni, R.B., Ross, J. (Eds.), Photon-Vegetation Interactions. Springer Berlin Heidelberg, Berlin, Heidelberg, pp. 139–159. https://doi.org/10.1007/978-3-642-75389-3_5

[18] Li, X., Strahler, A., 1985. Geometric-Optical Modeling of a Conifer Forest Canopy. IEEE Trans. Geosci. Remote Sensing GE-23, 705–721.

https://doi.org/10.1109/TGRS.1985.289389

[19] Li, X., Zhang, W., Wang, B., Qiu, H., Jin, M., Qi, P., 2025. OGAIS: OpenGL-Driven GPU Acceleration Methodology for 3D Hyperspectral Image Simulation. Remote Sensing 17, 1841. https://doi.org/10.3390/rs17111841

[20] Lucht, W., Schaaf, C.B., Strahler, A.H., 2000. An algorithm for the retrieval of albedo from space using semiempirical BRDF models. IEEE Trans. Geosci. Remote Sensing 38, 977–998. https://doi.org/10.1109/36.841980

[21] Ma, R., He, S., 2024. Hyperspectral Neural Radiance Field Method Based on Reference Spectrum. IEEE Access 12, 133018–133029. https://doi.org/10.1109/ACCESS.2024.3459917

[22] Meyer, L., Grün, J., Weiherer, M., Egger, B., Stamminger, M., Franke, L., 2026. Multi-Spectral Gaussian Splatting with Neural Color Representation. Computer Graphics Forum e70337. https://doi.org/10.1111/cgf.70337

[23] Narayanan, S.K., Zhao, L., Gan, L., Chen, Y., 2025. Hyperspectral Gaussian Splatting. https://doi.org/10.48550/arXiv.2505.21890

[24] Nicodemus, F.E., 1965. Directional Reflectance and Emissivity of an Opaque Surface. Appl. Opt. 4, 767. https://doi.org/10.1364/AO.4.000767

[25] Nicodemus, F.E., Richmond, J.C., Hsia, J.J., Ginsberg, I.W., Limperis, T., 1977. Geometrical considerations and nomenclature for reflectance (No. NBS MONO 160). National Bureau of Standards, Gaithersburg, MD. https://doi.org/10.6028/NBS.MONO.160

[26] Pinty, B., Verstraete, M.M., Dickinson, R.E., 1989. A physical model for predicting bidirectional reflectances over bare soil. Remote Sensing of Environment 27, 273–288. https://doi.org/10.1016/0034-4257(89)90088-6

[27] Qi, J., Jiang, J., Zhou, K., Xie, D., Huang, H., 2023. Fast and Accurate Simulation of Canopy Reflectance under Wavelength-Dependent Optical Properties Using a Semi-Empirical 3D Radiative Transfer Model. J Remote Sens 3, 0017. https://doi.org/10.34133/remotesensing.0017

[28] Qi, J., Xie, D., Yin, T., Yan, G., Gastellu-Etchegorry, J.-P., Li, L., Zhang, W., Mu, X., Norford, L.K., 2019. LESS: LargE-Scale remote sensing data and image simulation framework over heterogeneous 3D scenes. Remote Sensing of Environment 221, 695–706. https://doi.org/10.1016/j.rse.2018.11.036

[29] Román, M.O., Gatebe, C.K., Schaaf, C.B., Poudyal, R., Wang, Z., King, M.D., 2011. Variability in surface BRDF at different spatial scales (30m–500m) over a mixed agricultural landscape as retrieved from airborne and satellite spectral measurements. Remote Sensing of Environment 115, 2184–2203. https://doi.org/10.1016/j.rse.2011.04.012

[30] Ross, J., 1981. The radiation regime and architecture of plant stands. Springer Netherlands, Dordrecht. https://doi.org/10.1007/978-94-009-8647-3

[31] Roujean, J., Leroy, M., Deschamps, P., 1992. A bidirectional reflectance model of the Earth’s surface for the correction of remote sensing data. J. Geophys. Res. 97, 20455–20468. https://doi.org/10.1029/92JD01411

[32] Sandmeier, St., Müller, Ch., Hosgood, B., Andreoli, G., 1998. Physical Mechanisms in Hyperspectral BRDF Data of Grass and Watercress. Remote Sensing of Environment 66, 222–233. https://doi.org/10.1016/S0034-4257(98)00060-1

[33] Schaaf, C.B., Gao, F., Strahler, A.H., Lucht, W., Li, X., Tsang, T., Strugnell, N.C., Zhang, X., Jin, Y., Muller, J.-P., Lewis, P., Barnsley, M., Hobson, P.,

Disney, M., Roberts, G., Dunderdale, M., Doll, C., d'Entremont, R.P., Hu, B., Liang, S., Privette, J.L., Roy, D., 2002. First operational BRDF, albedo nadir reflectance products from MODIS. Remote Sensing of Environment 83, 135–148. https://doi.org/10.1016/S0034-4257(02)00091-3

[34] Schaepman-Strub, G., Schaepman, M.E., Painter, T.H., Dangel, S., Martonchik, J.V., 2006. Reflectance quantities in optical remote sensing—definitions and case studies. Remote Sensing of Environment 103, 27–42. https://doi.org/10.1016/j.rse.2006.03.002

[35] Schlick, C., 1994. An Inexpensive BRDF Model for Physically-based Rendering. Computer Graphics Forum 13, 233–246. https://doi.org/10.1111/1467-8659.1330233

[36] Schott, J.R., Brown, S.D., Raqueño, R.V., Gross, H.N., Robinson, G., 1999. An Advanced Synthetic Image Generation Model and its Application to Multi/Hyperspectral Algorithm Development. Canadian Journal of Remote Sensing 25, 99–111. https://doi.org/10.1080/07038992.1999.10874709

[37] Sinha, S.N., Graf, H., Weinmann, M., 2025. SpectralGaussians: Semantic, spectral 3D Gaussian splatting for multi-spectral scene representation, visualization and analysis. ISPRS Journal of Photogrammetry and Remote Sensing 227, 789–803. https://doi.org/10.1016/j.isprsjprs.2025.06.008

[38] Soenen, S.A., Peddle, D.R., Coburn, C.A., 2005. SCS+C: a modified Sun-canopy-sensor topographic correction in forested terrain. IEEE Trans. Geosci. Remote Sensing 43, 2148–2159. https://doi.org/10.1109/TGRS.2005.852480

[39] Stark, B., Zhao, T., Chen, Y., 2016. An analysis of the effect of the bidirectional reflectance distribution function on remote sensing imagery accuracy from Small

Unmanned Aircraft Systems, in: 2016 International Conference on Unmanned Aircraft Systems (ICUAS). Presented at the 2016 International Conference on Unmanned Aircraft Systems (ICUAS), IEEE, Arlington, VA, USA, pp. 1342–1350. https://doi.org/10.1109/ICUAS.2016.7502566

[40] Thirgood, C., Mendez, O., Ling, E., Storey, J., Hadfield, S., 2025. HyperGS: Hyperspectral 3D Gaussian Splatting, in: 2025 IEEE/CVF Conference on Computer Vision and Pattern Recognition (CVPR). Presented at the 2025 IEEE/CVF Conference on Computer Vision and Pattern Recognition (CVPR), IEEE, Nashville, TN, USA, pp. 5970–5979. https://doi.org/10.1109/CVPR52734.2025.00560

[41] Verhoef, W., 1984. Light scattering by leaf layers with application to canopy reflectance modeling: The SAIL model. Remote Sensing of Environment 16, 125–141. https://doi.org/10.1016/0034-4257(84)90057-9

[42] Walter, B., Marschner, S.R., Li, H., Torrance, K.E., 2007. Microfacet Models for Refraction through Rough Surfaces. Rendering Techniques. https://doi.org/10.2312/EGWR/EGSR07/195-206

[43] Wang, Y., Kallel, A., Yang, X., Regaieg, O., Lauret, N., Guilleux, J., Chavanon, E., Gastellu-Etchegorry, J.-P., 2022. DART-Lux: An unbiased and rapid Monte Carlo radiative transfer method for simulating remote sensing images. Remote Sensing of Environment 274, 112973. https://doi.org/10.1016/j.rse.2022.112973

[44] Wanner, W., Li, X., Strahler, A.H., 1995. On the derivation of kernels for kernel-driven models of bidirectional reflectance. J. Geophys. Res. 100, 21077–21089. https://doi.org/10.1029/95JD02371

[45] Wu, S., Wen, J., Xiao, Q., Liu, Q., Hao, D., Lin, X., You, D., 2019. Derivation of

Kernel Functions for Kernel-Driven Reflectance Model Over Sloping Terrain. IEEE J. Sel. Top. Appl. Earth Observations Remote Sensing 12, 396–409. https://doi.org/10.1109/JSTARS.2018.2854771

[46] Yan, K., Li, H., Song, W., Tong, Y., Hao, D., Zeng, Y., Mu, X., Yan, G., Fang, Y., Myneni, R.B., Schaaf, C., 2022. Extending a Linear Kernel-Driven BRDF Model to Realistically Simulate Reflectance Anisotropy Over Rugged Terrain. IEEE Trans. Geosci. Remote Sensing 60, 1–16. https://doi.org/10.1109/TGRS.2021.3064018

[47] Yang, Z., Gao, X., Sun, Y.-T., Huang, Y.-H., Lyu, X., Zhou, W., Jiao, S., Qi, X., Jin, X., 2024. Spec-Gaussian: Anisotropic View-Dependent Appearance for 3D Gaussian Splatting, in: Advances in Neural Information Processing Systems 37. Presented at the Advances in Neural Information Processing Systems 37, Neural Information Processing Systems Foundation, Inc. (NeurIPS), Vancouver, BC, Canada, pp. 61192–61216. https://doi.org/10.52202/079017-1956

[48] Yao, Y., Zeng, Z., Gu, C., Zhu, X., Zhang, L., 2025. Reflective Gaussian Splatting, in: Yue, Y., Garg, A., Peng, N., Sha, F., Yu, R. (Eds.), International Conference on Learning Representations. pp. 68695–68711.

[49] Ye, K., Hou, Q., Zhou, K., 2024. 3D Gaussian Splatting with Deferred Reflection, in: Special Interest Group on Computer Graphics and Interactive Techniques Conference Conference Papers 24. Presented at the SIGGRAPH ’24: Special Interest Group on Computer Graphics and Interactive Techniques Conference, ACM, Denver CO USA, pp. 1–10. https://doi.org/10.1145/3641519.3657456

[50] Zhang, W., Tang, J., Zhang, Weiqi, Fang, Y., Liu, Y.-S., Han, Z., 2025b. MaterialRefGS: Reflective Gaussian Splatting with Multi-view Consistent

Material Inference, in: Advances in Neural Information Processing Systems 38. Presented at the Advances in Neural Information Processing Systems 38, Neural Information Processing Systems Foundation, Inc. (NeurIPS), San Diego, California, USA and Mexico City, Mexico, pp. 99861–99882. https://doi.org/10.52202/085713-3007

[51] Zhang, Y., Chen, A., Wan, Y., Song, Z., Yu, J., Luo, Y., Yang, W., 2025a. Ref-GS: Directional Factorization for 2D Gaussian Splatting, in: 2025 IEEE/CVF Conference on Computer Vision and Pattern Recognition (CVPR). Presented at the 2025 IEEE/CVF Conference on Computer Vision and Pattern Recognition (CVPR), IEEE, Nashville, TN, USA, pp. 26483–26492. https://doi.org/10.1109/CVPR52734.2025.02466

[52] Xu, J., Zhao, Y., Long, F., 2000. Analysis and application of directional reflectance in satellite remotely sensed data. Journal of Remote Sensing 4 (S1), 110. https://doi.org/10.11834/jrs.2000S117

[53] Liu, Q., Tang, Y., Li, J., Du, Y., Wen, J., Yao, Y., Huang, H., Tian, G., 2009. Research progress on modeling and inversion of remote sensing radiative transfer. Journal of Remote Sensing 13 (S1), 182. https://doi.org/10.11834/jrs.20090023